\documentclass[journal]{IEEEtran}
\usepackage{graphicx} 
\usepackage{color}
\usepackage{colortbl} 
\usepackage{dcolumn}
\usepackage[table]{xcolor}
\usepackage{booktabs}
\usepackage{multirow}
\usepackage{pifont}
\usepackage[export]{adjustbox}
\usepackage{epstopdf}
\usepackage{subfigure}
\usepackage{amsfonts}  
\usepackage{caption}
\usepackage{amsmath}
\usepackage{algorithmic}
\usepackage{array}
\usepackage{stfloats}
\usepackage{url}

\usepackage{tabularx,booktabs}
\usepackage[linesnumbered,lined,ruled]{algorithm2e}
\usepackage{threeparttable}
\usepackage{rotating}
\usepackage{footmisc}
\usepackage{float}
\usepackage{makecell}
\usepackage[switch]{lineno}
\usepackage{algorithmic}

\makeatletter 
\newcommand{\removelatexerror}{\let\@latex@error\@gobble}
\makeatother

\usepackage{verbatim}

\begin{document}
\captionsetup[figure]{labelformat={default},labelsep=period,name={Fig.},labelfont={small},  singlelinecheck=off, textfont={small}} 
\captionsetup[table]{format=plain,labelformat=simple,justification=centering, labelsep=newline, singlelinecheck=false,labelfont={small}, textfont={sc, small}}

\title{Local-to-Global Cross-Modal Attention-Aware Fusion for HSI-X Semantic Segmentation}

\author{Xuming~Zhang,~\IEEEmembership{} 
        Naoto Yokoya,~\IEEEmembership{}
        Xingfa~Gu,~\IEEEmembership{} 
        Qingjiu~Tian,~\IEEEmembership{}
        and Lorenzo Bruzzone,~\IEEEmembership{Fellow,~IEEE}%
            
\thanks{
This work was supported in part by the National Key Research and Development Program of China under Grant 2023YFF1303903, in part by the Open Fund of State Key Laboratory of Urban and Regional Ecology under Grant SKLURE2023-2-6, and in part by the National Natural Science Foundation of China under Grant 42101321. (Corresponding author: Qingjiu Tian).

Xuming Zhang and Qingjiu Tian are with the International Institute for Earth System Science, Nanjing University, Nanjing 210023, China (e-mail: xuming.zhang.rs@gmail.com, tianqj@nju.edu.cn).

Naoto Yokoya is with the Department of Complexity Science and Engineering, Graduate School of Frontier Sciences, The University of Tokyo, Chiba 277-8561, Japan, and also with the RIKEN Center for Advanced Intelligence Project, Tokyo 103-0027, Japan (e-mail: yokoya@k.u-tokyo.ac.jp).

Xingfa Gu is with the Key Laboratory of Computational Optical Imaging Technology, Aerospace Information Research Institute, Chinese Academy of Sciences, Beijing 100094, China (e-mail: gaolr@aircas.ac.cn).

Lorenzo Bruzzone is with the Department of Information Engineering and Computer Science, University of Trento, 38050 Trento, Italy (e-mail: lorenzo.bruzzone@ing.unitn.it).
}}

\markboth{JOURNAL OF LATEX CLASS FILES}%
{Shell \MakeLowercase{\textit{et al.}}: Bare Demo of IEEEtran.cls for Journals}

\maketitle

\begin{abstract}
Hyperspectral image (HSI) classification has recently reached its performance bottleneck. Multimodal data fusion is emerging as a promising approach to overcome this bottleneck by providing rich complementary information from the supplementary modality (X-modality). However, achieving comprehensive cross-modal interaction and fusion that can be generalized across different sensing modalities is challenging due to the disparity in imaging sensors, resolution, and content of different modalities. In this study, we propose a Local-to-Global Cross-modal Attention-aware Fusion (LoGoCAF) framework for HSI-X classification that jointly considers efficiency, accuracy, and generalizability. LoGoCAF adopts a pixel-to-pixel two-branch semantic segmentation architecture to learn information from HSI and X modalities. The pipeline of LoGoCAF consists of a local-to-global encoder and a lightweight multilayer perceptron (MLP) decoder. In the encoder, convolutions are used to encode local and high-resolution fine details in shallow layers, while transformers are used to integrate global and low-resolution coarse features in deeper layers. The MLP decoder aggregates information from the encoder for feature fusion and prediction. In particular, two cross-modality modules, the feature enhancement module (FEM) and the feature interaction and fusion module (FIFM), are introduced in each encoder stage. The FEM is used to enhance complementary information by combining the feature from the other modality across direction-aware, position-sensitive, and channel-wise dimensions. With the enhanced features, the FIFM is designed to promote cross-modality information interaction and fusion for the final semantic prediction. Extensive experiments demonstrate that our LoGoCAF achieves superior performance and generalizes well on various multimodal fusion and classification tasks. The code will be made publicly available.
\end{abstract}

\begin{IEEEkeywords}
Hyperspectral image (HSI), cross-modal fusion, classification, semantic segmentation, deep learning, transformer, remote sensing.
\end{IEEEkeywords}

\IEEEpeerreviewmaketitle
\section{Introduction}
Hyperspectral images (HSIs) contain abundant spectral information that allows to characterize each material with a distinct signature, similar to a unique fingerprint, facilitating its identification \cite{8113122, my2}. HSI classification, or semantic segmentation, provides a plausible solution for transforming an HSI image into its underlying semantically meaningful regions, enabling pixel-wise dense scene understanding in many applications \cite{10231003}, including land-use analysis \cite{HONG202168}, environmental monitoring \cite{r1}, and urban development observation \cite{r3}. Despite significant advances in HSI classification, existing approaches often struggle to achieve high-quality, fine-grained identification in complex scenes, such as when different categories share similar materials.

With the advancement of remote sensing imaging techniques, it is possible to simultaneously acquire data with multiple modalities to monitor the same geographic area \cite{7740215}. Different modalities can provide HSIs with complementary signals. For instance, digital surface model (DSM) captures elevation information, light detection and ranging (LiDAR) provides accurate 3D spatial information and synthetic aperture radar (SAR) offers structural information \cite{HONG202168}. Therefore, fusion of multimodal data is essential to achieve accurate and reliable classification results.

Existing multimodal data fusion approaches fall into three broad categories: pixel-level fusion, feature-level fusion, and decision-level fusion. Pixel-level fusion \cite{LI2017100} usually deploys a single technique to jointly extract features from HSI and another modality, with fusion occurring at the input stage. However, these methods are typically tailored for specific modalities and lack scalability \cite{10231003}. Feature-level fusion \cite{6891148} employs two algorithms to extract features from the HSI and another modality, respectively. The extracted features are then fused into a unified feature vector for prediction. Decision-level fusion methods, such as the simple majority voting \cite{618255} and weighted averaging \cite{Wei_aver}, aggregate independent classifications from each modality \cite{763301}. Among these three fusion approaches, feature-level fusion is more scalable and has gained increasing attention \cite{10231003}.

Considerable progress has been made in feature-level fusion of multimodal data. Various traditional methods, including extended attribute profiles (EAPs) \cite{6237479}, local binary patterns \cite{8718389}, and manifold learning \cite{8718389}, have been developed to focus on feature extraction from multimodal data. In \cite{6237479}, EAPs are used to extract HSI and LiDAR features, which are then concatenated and classified using random forest \cite{r5} and support vector machine (SVM) \cite{r4} classifiers. The fusion technique was improved in \cite{8000656} by incorporating sparse and low-rank techniques into the extinction profiles of HSI and LiDAR data. The shared and specific feature learning (S2FL) model \cite{8718389} gathers information from HSI and another modality by aligning shared components between multimodal sources on manifolds while segregating their specific information. Yokoya et al. \cite{7946218} presented a comparative review of HSI and higher spatial resolution multispectral image (MSI) fusion techniques, including Bayesian, hypersharpening and unmixing based methods, and provided directions for the future development of HSI–MSI fusion. Although these traditional fusion classification techniques have provided valuable solutions and insights, they still exhibit drawbacks, such as the requirement for expert feature engineering and the challenges of fusing heterogeneous data sources.

Deep learning (DL), as an automatic feature learning technique, has demonstrated outstanding capabilities in feature extraction and has been widely applied to multimodal data fusion \cite{10231003, 9772757, 9150738}. Among DL algorithms, convolutional neural networks (CNNs) are the go-to models for feature extraction due to their efficiency and ease of optimization \cite{rCCT}. EndNet, a fully connected (FC) network \cite{9179756}, deploys two FC modules for feature extraction and one for feature fusion and classification. Some approaches \cite{10147800, 9179756} follow a two-branch architecture for feature extraction, with each branch corresponding to one modality. Alternatively, three-stream architectures \cite{8068943, 9813366} use a 1-D CNN and a 2-D CNN separately for spectral and spatial feature learning from HSI data, and another 2-D CNN for spatial feature learning from another modality. 

However, CNNs treat each pixel and band as equally important, which is impractical as different bands and pixels contribute differently to feature extraction and identification \cite{my3}. To overcome this drawback, attention mechanisms have been developed to guide models on ``what'' and “where" to focus. Many works have incorporated attention mechanisms to refine feature maps, mitigating the equal treatment of convolution kernels. In \cite{9813366, rCALC}, attention blocks are applied to a single modality, highlighting only unimodal features. Subsequently, some architectures \cite{9150738, 9790044, WANG20221} generate attention masks from one modality to enhance representations of the other. For example, FusAtNet \cite{9150738} develops a ``cross-attention'' module to exploit LiDAR-derived attention maps to highlight spatial features of HSIs. A self- and cross-guided attention model \cite{9790044} refines information from HSI and LiDAR branches while providing cross-guidance from the LiDAR branch to the HSI branch. Unlike the common practice of simply concatenating individually extracted features, Xue et al. \cite{9526908} designed an attentional feature fusion module that adaptively and dynamically fuses information from multimodal data. The multistage information complementary fusion network (MCFNet) \cite{10216780} provides comprehensive multimodal information interaction and complementation for HSI and X-modal (e.g., multispectral, SAR, and LiDAR) data.

Recently, vision transformer (ViT) \cite{ViT} and its follow-ups \cite{rSwinViT, rHVT, rCCT, rMobileViT} have achieved leading performance in various visual tasks, such as object detection, semantic segmentation, and image classification. ViT replaces the inductive biases toward local modeling inherent in convolutions with global modeling driven by self-attention mechanisms, demonstrating that transformers can surpass CNNs with large-scale training data. Self-attention, a core building block of ViT, enables adaptive information aggregation and long-range modeling. Several studies have explored the application of transformers to multimodal image processing and analysis. The multimodal fusion transformer (MFT) \cite{rMFT} is a ViT-based network in which the input patches are taken from HSIs, while the class token is generated from LiDAR data. The deep hierarchical ViT presented in \cite{9755059} consists of spectral, spatial, and LiDAR branches, where the spectral branch adopts a spectral transformer to handle the long-range spectral dependencies in HSIs, while the spatial and LiDAR branches employ spatial  transformers to learn spatial information from HSI and LiDAR data, respectively. Additionally, a cross-attention feature fusion module is designed to adaptively and dynamically fuse heterogeneous features from multimodal data. Similarly, Zhang et al. \cite{10145469} proposed a multimodal transformer network using multiple transformers to capture individual and shared features from two modalities. However, these transformer networks lack the inductive biases that CNN possesses and therefore require large amounts of data and computation to compensate. 

To alleviate the shortcomings of transformers, a promising solution is to incorporate the inductive biases of CNNs into transformers. From this perspective, hybrid architectures combining convolutions and transformers have emerged. The global–local transformer network (GLT-Net) \cite{9926173} employs CNN and transformer modules for multiscale local spatial feature learning and global spectral feature learning, respectively. Zhao et al. \cite{rFusionHCT} developed a dual-branch hierarchical network, called Fusion\_HCT, that sequentially performs feature extraction with CNN and transformer blocks, followed by a cross-token attention fusion structure for feature fusion. The local information interaction transformer (LIIT) \cite{10004002} model employs a dual-branch transformer to exploit sequence information from multimodal data, which is then fused and fed into a convolutional transformer module for further classification.

Although the aforementioned studies have achieved a certain degree of success in multimodal data fusion and classification from different perspectives, they still exhibit relevant limitations:

1) Current multimodal data fusion and classification models primarily follow the patch-based learning pipeline, which produces the category of a pixel by learning information from a small spatial patch centered on that pixel. This scheme leads to redundant computation due to significant overlap between adjacent patches, and the limited patch size hampers long-range spatial dependence modeling \cite{my3}. Fusion-FCN \cite{rFusionFCN}, the first-place winner of the 2018 IEEE Data Fusion Contest (DFC), is based on a fully convolutional network (FCN) \cite{7478072}, demonstrating the great potential of FCNs in data fusion and classification. However, Fusion-FCN is a pure CNN lacking global connectivity. 

2) Some approaches \cite{9179756, rFusionFCN, 9813366} use a simple multimodal fusion strategy, such as the common concatenation operator, which lacks generalization across different multimodal data. Other approaches \cite{9150738, 10216780} employ self-attention based cross-modality fusion models. However, the computational burden imposed by these self-attention mechanisms is not affordable for networks with high-resolution input images due to the high computational cost and insufficient training data. Therefore, designing a high-efficiency, high-performance, and high-generalization cross-modality fusion model for multimodal data fusion remains an open problem.

\begin{figure*}[htbp] 
\centerline{\includegraphics[width=18cm]{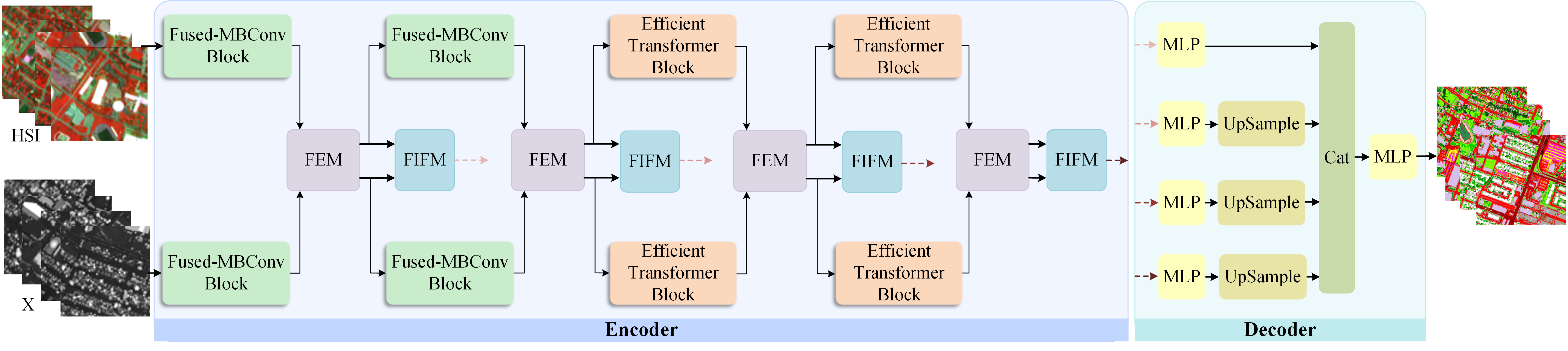}} 
\caption{Overview illustration of the proposed LoGoCAF framework for HSI-X classification. Here, FEM and FIFM represents the proposed feature enhancement module and feature interaction and 
fusion module, respectively, and ``Cat'' is the concatenation operation.}
\label{Overview} 
\end{figure*}

To overcome the aforementioned limitations, we propose a \textbf{Lo}cal-to-\textbf{G}l\textbf{o}bal \textbf{C}ross-modal \textbf{A}ttention-aware \textbf{F}usion (LoGoCAF) framework for HSI-X classification that jointly considers efficiency, accuracy, and versatility. LoGoCAF is built as a pixel-to-pixel two-branch semantic segmentation architecture (HSI- and X-modal branch). The pipeline of LoGoCAF comprises a local-to-global encoder and a lightweight multilayer perceptron (MLP) decoder. In the encoder, convolution is employed in shallow layers to encode local and high-resolution fine details, while transformers integrate global and low-resolution coarse features in deep layers. The MLP decoder aggregates information from the encoder for feature fusion and prediction. Furthermore, two cross-modality modules, the feature enhancement module (FEM), and feature interaction and fusion module (FIFM), are designed to promote multimodal feature enhancement, interaction and fusion. The FEM enhances bi-modal features by leveraging direction-aware, position-sensitive, and channel-wise correlations. This allows both modalities to emphasize complementary informative cues from each other and alleviates the effects of uncertainties and noisy measurements in each modality. Additionally, FIFM uses an efficient cross-attention mechanism to exploit the most relevant regions in one modality and generate an attention mask that highlights the features of the other modality in an adaptive and content-aware manner.

The main contributions of this study are as follows.

1) Extending transformer-based pixel-to-pixel semantic segmentation research to HSI-X fusion and classification by introducing LoGoCAF, which jointly considers efficiency, accuracy, and generalizability. LoGoCAF is specifically tailored to extract discriminant features from diverse modalities, making it a valuable contribution to HSI-X image fusion and classification.

2) Developing a two-branch semantic segmentation pipeline (RGB- and X-modal branch), which consists of a hybrid hierarchical encoder and a lightweight MLP decoder. The encoder combines the strengths of CNNs and transformers to model local-to-global and high-to-low-resolution information efficiently. The decoder produces powerful representations without relying on complex and computationally demanding modules.

3) Designing two cross-modality modules, FEM and FIFM, to enhance multimodal features extraction, interaction and fusion, further promoting performance.

The remainder of this paper is organized as follows. In Section \uppercase\expandafter{\romannumeral2}, we provide a detailed description of the proposed LoGoCAF framework. Section \uppercase\expandafter{\romannumeral3} presents and discusses the experiments and results, while Section \uppercase\expandafter{\romannumeral4} conducts a network analysis. Finally, Section \uppercase\expandafter{\romannumeral5} draws concluding remarks.

\section{Methodology}
An overview of the proposed LoGoCAF framework is shown in Fig.~\ref{Overview}. It consists of a hybrid encoder and a lightweight decoder. The encoder uses two parallel and interacting backbones to extract features from HSI- and X-modal (e.g., LiDAR, DEM, SAR) data. It follows a multi-stage hierarchical structure to generate CNN-like multi-level features, where four stages are deployed. The first two stages employ the Fused-MBConv block \cite{rEfficientnetV2} to encode local and high-resolution detailed information. The subsequent two stages incorporate the efficient transformer block \cite{rSegFormer} to integrate global and low-resolution coarse features. The placement of convolutions before transformers is based on the prior knowledge that convolutions are good at encoding local information in shallow layers due to their strong inductive bias. To enhance the complementary information of both modalities, the FEM is strategically designed and inserted between two adjacent stages. The output enhanced features are sent to the next stage for further information learning, while being sent to the FIFM for feature interaction, fusion, and subsequent classification. Next, a decoder is employed to fuse the multi-level feature maps to produce the final classification map. Details of the Fused-MBConv blocks, efficient transformer block, FEM, FIFM, and decoder are presented in the following sections.

\subsection{Fused-MBConv} 
\begin{figure}[th]
 \centering
\subfigure[]
{   
\begin{minipage}{2.65cm}
\includegraphics[width=2.65cm]{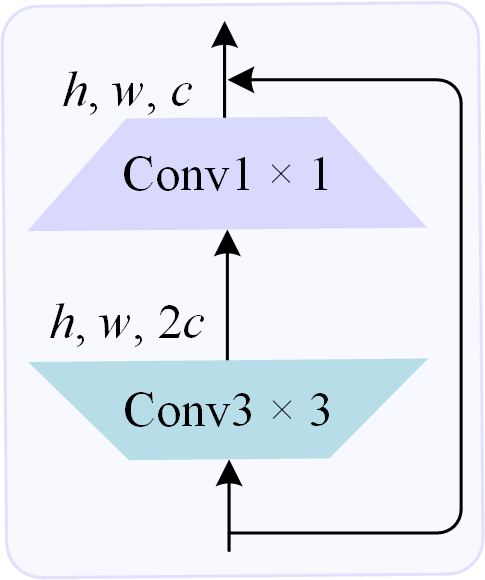}
\end{minipage}
}
\subfigure[]
{
\begin{minipage}{5.55cm}
\centering
\includegraphics[width=5.55cm]{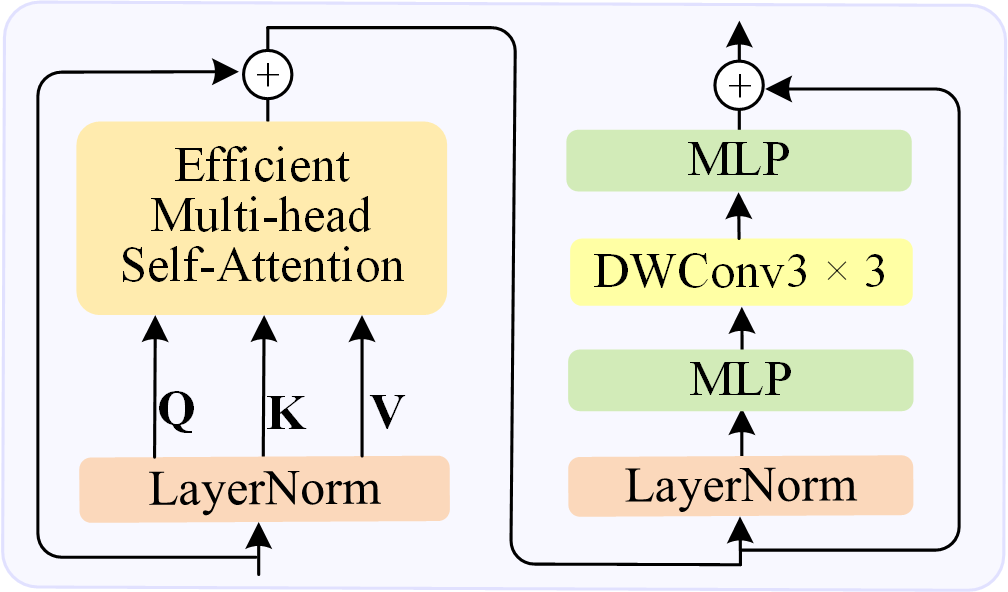}
\end{minipage}
}
\caption{ Structure of (a) Fused-MBConv without SE and (b) Efficient transformer block.}
\label{Block} 
\end{figure}

The Fused-MBConv block adopted in LoGoCAF, as illustrated in Fig.~\ref{Block}(a), first uses a $3 \times 3$ expansion convolution to capture local information while expanding the channels. Following this, a $1 \times 1$ convolution reduces the channel dimension back to its original dimension, so that the input and output can be added. The expansion ratio of the $3\times 3$ expansion convolution is set to 2. Moreover, we also discuss different convolution blocks (i.e., Fused-MBConv with Squeeze-and-Excitation (SE) and MBConv) in experiments and finally adopt the Fused-MBConv without SE. Incorporating the Fused-MBConv block in the early stages is more effective and produces richer tokens for subsequent transformers.

\subsection{ Efficient Transformer Block} 
Within transformer architectures, the main computational bottleneck is the self-attention mechanism. In its foundational formulation, the query $\mathbf{Q}\in {{\mathbb{R}}^{N\times C}}$, key $\mathbf{K}\in {{\mathbb{R}}^{N\times C}}$ and value $\mathbf{V}\in {{\mathbb{R}}^{N\times C}}$ are linearly transformed from the input $\mathbf{X}\in {{\mathbb{R}}^{H\times W\times C}}$, where $N=H\times W$, and $H$, $W$ and $C$ represent the length, width and the number of bands of $\mathbf{X}$, respectively. The transformation from  $H\times W\times C$ to  $N\times C $ is omitted for simplicity. The self-attention module is then expressed as follows:
\begin{equation}
\text{Attention}\left( \mathbf{Q},\mathbf{K},\mathbf{V} \right)=\text{Softmax}(\mathbf{Q}{{\mathbf{K}}^{T}}/\sqrt{d})\mathbf{V}.
\end{equation}

The computational complexity of this operation is $O\left( {{N}^{2}} \right)$, posing challenges particularly for high-resolution images. To mitigate this problem, efficient self-attention \cite{10231003} is used to reduce the sequence length of K before the attention operation as follows:
\begin{equation}
{{\mathbf{K}}^{'}}=\text{Reshape}\left( \mathbf{K} \right)\in {{\mathbb{R}}^{\tfrac{N}{R}\times \left( C\cdot R \right)}},
\end{equation}
\begin{equation}
\mathbf{K}=\text{MLP}\left( {{\mathbf{K}}^{'}} \right)\in {{\mathbb{R}}^{\tfrac{N}{R}\times C}},
\end{equation}
where $\mathbf{K}\in {{\mathbb{R}}^{N\times C}}$ is first reshaped to ${{\mathbf{K}}^{'}}\in {{\mathbb{R}}^{\tfrac{N}{R}\times \left( C\cdot R \right)}}$, and then an MLP is employed to transform ${{\mathbf{K}}^{'}}$ to $\mathbf{K}\in {{\mathbb{R}}^{\tfrac{N}{R}\times C}}$. Consequently, the complexity of the self-attention mechanism is reduced from $O\left( {{N}^{2}} \right)$ to $O\left( \tfrac{{{N}^{2}}}{R} \right)$. The efficient multi-head self-attention (EMSA) module is defined by considering $h$ ``heads'', i.e., the input $\mathbf{X}$ is equally divided into $h$ ``heads'' by channel, and then the efficient self-attention function is applied to each ``heads''.

The original feed-forward network (FFN) in ViT \cite{ViT} consists of two MLP layers separated by a GELU activation function:
\begin{equation}
\text{FFN= MLP (GELU}\left( \text{MLP}\left( {{\mathbf{X}}_{\text{in}}} \right) \right).
\end{equation}

Previous research \cite{rCMT,rEfficientViT,rUniNeXt} has demonstrated that embedding a ${3\times3}$ depth-wise convolution (DWConv) into the FFN can provide positional information for transformer layers and improve performance. Our efficient transformer block inherits this DWConv embedding mechanism in the FFN (denoted as DWFFN). As shown in Fig.~\ref{Block}(b), the DWFFN consists of two MLP layers separated by a ${3\times3}$ DWConv with a GELU activation function, expressed as:
\begin{equation}
\text{DWFFN= MLP (GELU}\left( \text{DWConv}\left( \text{MLP}\left( {{\mathbf{X}}_{\text{in}}} \right) \right) \right),
\end{equation}
where ${{\mathbf{X}}_{\text{in}}}$ represents the outputs of the self-attention module, the first MLP layer expands the channel dimension by a factor of 2, and the second MLP layer reduces the dimension by the same ratio. 
With the above two components, the transformer block can be formulated as:
\begin{equation}
{{\mathbf{X}}^{'}}=\text{EMSA(LN(}\mathbf{X}\text{))+}\mathbf{X},
\end{equation}
\begin{equation}
\mathbf{Y}=\text{DWFFN(LN(}{{\mathbf{X}}^{'}}\text{))+}{{\mathbf{X}}^{'}}.
\end{equation}

\subsection{ Feature Enhancement Module (FEM)}
\begin{figure*}[tbp] 
\centerline{\includegraphics[width=18cm]{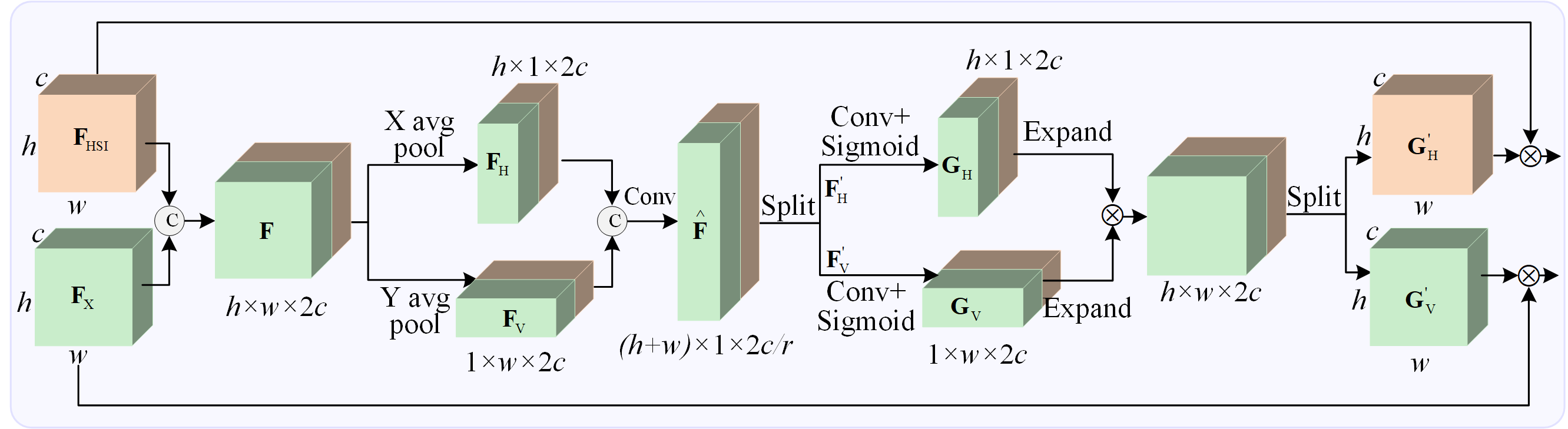}} 
\caption{Illustration of the proposed cross-modality feature enhancement module (FEM).}
\label{FEM} 
\end{figure*}

Information derived from different modalities typically provides complementary information, but may also introduce noise or destroyed information. Leveraging features from another modality can enhance complementary information, filter out noise, and restore destroyed information \cite{10231003,rRGBD}. Therefore, we propose a cross-modality FEM to enhance and recalibrate the modality-specific features of each modality at each stage of the encoder. The FEM processes the input features by considering not only the inter-channel relationships, but also directional and positional information.

Specifically, as shown in Fig.~\ref{FEM}, given the bimodal inputs ${{\mathbf{F}}_{\text{HSI}}}\in {{\mathbb{R}}^{h\times w\times c}}$ and ${{\mathbf{F}}_{\text{X}}}\in {{\mathbb{R}}^{h\times w\times c}}$, we concatenate them along the channel dimension to obtain $\mathbf{F}\in {{\mathbb{R}}^{h\times w\times 2c}}$. Then, a horizontal pooling $(h, 1)$ and a vertical pooling $(1, w)$ are applied to the concatenated features $\mathbf{F}$. The output of the horizontal pooling ${{\mathbf{F}}_{\text{H}}}\in {{\mathbb{R}}^{h\times 1\times 2c}}$ and vertical pooling ${{\mathbf{F}}_{\text{V}}}\in {{\mathbb{R}}^{1\times w\times 2c}}$ can be written as:
\begin{equation}
{{\mathbf{F}}_{\text{H}}}=\tfrac{1}{w}\sum\limits_{0\le j\le w}{{{\mathbf{F}}^{c}}}\left( h,j,2c \right),
\end{equation}
\begin{equation}
{{\mathbf{F}}_{\text{V}}}=\tfrac{1}{h}\sum\limits_{0\le i\le h}{\mathbf{F}}\left( i,w,2c \right).
\end{equation}   
After obtaining the aggregated features ${{\mathbf{F}}_{\text{H}}}$ and ${{\mathbf{F}}_{\text{V}}}$, we concatenate them into $\hat{\textbf{F}}\,\in {{\mathbb{R}}^{\left( h+w \right)\times 1\times 2c}}$, which are then fed into a $1\times 1$ convolution for dimensionality reduction, yielding:
\begin{equation}
\hat{\textbf{F}}\,=\delta \left( {{f}_{1}}\left( \left[ {{\mathbf{F}}_{\text{H}}},{{\mathbf{F}}_{\text{V}}} \right] \right) \right),
\end{equation}
where $\left[ \cdot ,\cdot  \right]$ denotes the concatenation operation along the spatial dimension, ${{f}_{1}}\left( \cdot  \right)$ represents the $1\times 1$ convolution and $\delta$ is an activation function. $\hat{\textbf{F}}\,\in {{\mathbb{R}}^{\left( h+w \right)\times 1\times 2c/r}}$ are the intermediate features that encode spatial information in both horizontal and vertical directions, where $r$ is the reduction ratio for controlling the block size. $\hat{\textbf{F}}\,$ are then split into $\mathbf{F}_{\text{H}}^{'}\in {{\mathbb{R}}^{h\times 1\times 2c/r}}$ and $\mathbf{F}_{\text{V}}^{'}\in {{\mathbb{R}}^{1\times w\times 2c/r}}$ along the spatial dimension :
\begin{equation}
\mathbf{F}_{\text{H}}^{'},\mathbf{F}_{\text{V}}^{'}={{F}_{\text{split}}}\left( \hat{\textbf{F}}\, \right).
\end{equation}
Next, another two $1\times 1$ convolutional transformations ${{f}_{\text{h}}}$ and ${{f}_{\text{w}}}$, followed by a sigmoid function, are separately utilized to transform $\mathbf{F}_{\text{H}}^{'}$ and $\mathbf{F}_{\text{V}}^{'}$ into tensors with the same channels of $2c$, yielding:
\begin{equation}
{{\mathbf{G}}_{\text{H}}}=\sigma  \left( {{f}_{\text{h}}}\left( \mathbf{F}_{\text{H}}^{'} \right) \right),
\end{equation}
\begin{equation}
{{\mathbf{G}}_{\text{V}}}=\sigma \left( {{f}_{\text{h}}}\left( \mathbf{F}_{\text{H}}^{'} \right) \right),
\end{equation}
where $\sigma $ represents the sigmoid function. The outputs ${{\mathbf{G}}_{\text{H}}}$ and ${{\mathbf{G}}_{\text{V}}}$ are expanded into the shape of $h\times w\times 2c$. These expanded outputs are multiplied as attention weights, which are then split along the channel dimension to generate the HSI and X-model attention maps: 
\begin{equation}
\mathbf{G}_{\text{HSI}}^{'},\mathbf{G}_{\text{X}}^{'}={{F}_{\text{split}}}\left( \text{expand}\left( {{\mathbf{G}}_{\text{H}}} \right)\otimes \text{expand}\left( {{\mathbf{G}}_{\text{V}}} \right) \right),
\end{equation}
where $\otimes $ refers to the element-wise multiplication and ${{F}_{\text{split}}}$ represents the split operation. Finally, the output features after cross-modality feature enhancement and rectification can be written as follows:
\begin{equation}
\text{HS}{{\text{I}}_{\text{rec}}}={{\mathbf{F}}_{\text{HSI}}}\otimes \mathbf{G}_{\text{HSI}}^{'},
\end{equation}
\begin{equation}
{{\text{X}}_{\text{rec}}}={{\mathbf{F}}_{\text{X}}}\otimes \mathbf{G}_{\text{X}}^{'}.
\end{equation}

By incorporating horizontal and vertical strip pooling operations, FEM can easily establish long-range dependencies between discretely distributed regions. In addition, FEM can identify the strengths of each modality and combine the most informative cross-modality features into an efficient representation. 

\begin{figure*}[tbp] 
\centerline{\includegraphics[width=17cm]{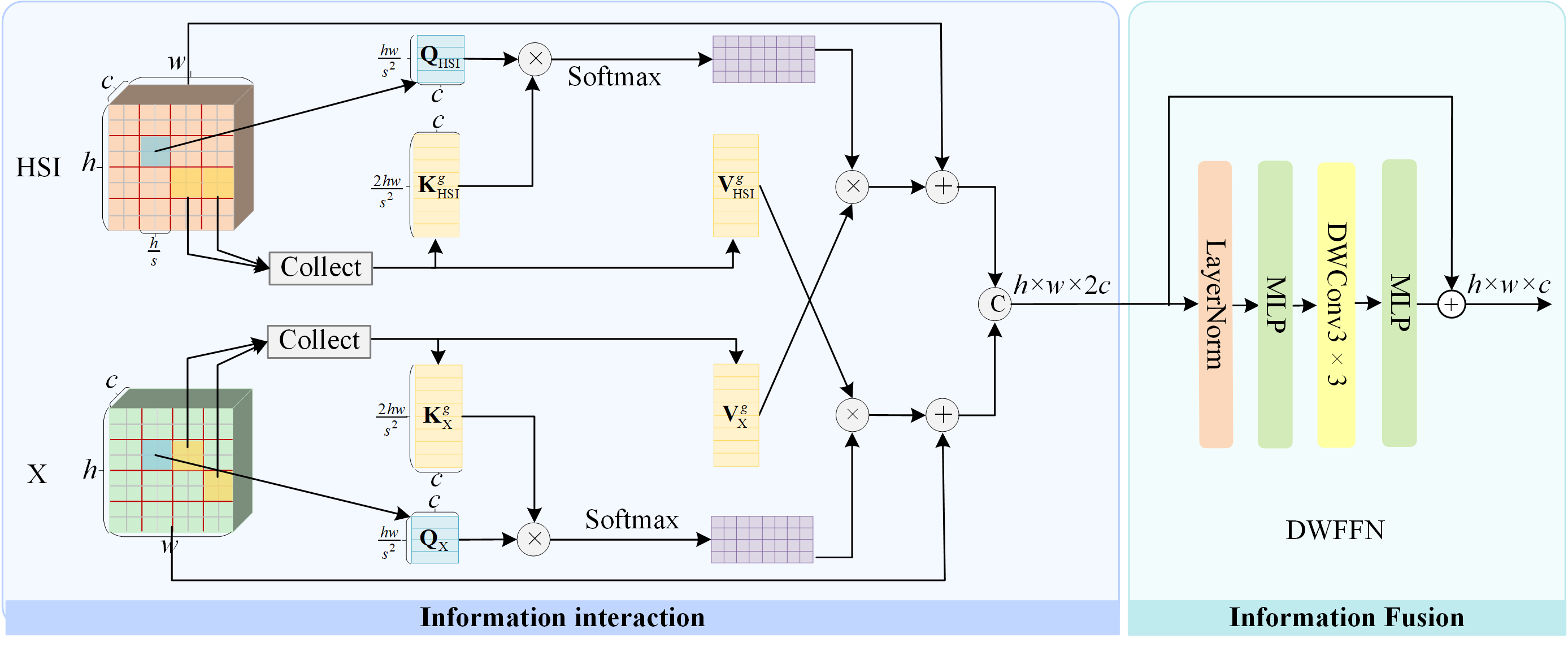}} 
\caption{Illustration of the proposed feature interaction and fusion module (FIFM).}
\label{FIFM} 
\end{figure*}

\subsection{Feature Interaction and Fusion Module (FIFM)}

After obtaining the enhanced modality-specific features at each stage, the FIFM is constructed to promote cross-modality information interaction and fusion. As illustrated in  Fig.~\ref{FIFM}, in the information interaction stage, a cross-attention mechanism is designed to exchange information between two modalities globally. The exchanged information is then concatenated and transformed to the original size via the DWFNN block in the fusion stage. Details of the cross-attention mechanism are presented below.

The main idea of the cross-attention mechanism is to select the most relevant key-value pairs at a coarse region level, preserving only the select subsets for further processing. Subsequently, a fine-grained token-to-token attention is applied to the selected regions. As shown in Fig.~\ref{FIFM}, given the bi-modal input feature maps ${{\mathbf{F}}_{\text{HSI}}}\in {{\mathbb{R}}^{h\times w\times c}}$ and ${{\mathbf{F}}_{\text{X}}}\in {{\mathbb{R}}^{h\times w\times c}}$, they are divided into $s\times s$ non-overlapping regions, each containing $hw/s^{2}$ feature vectors. This step is accomplished by reshaping ${{\mathbf{F}}_{\text{HSI}}}$ and ${{\mathbf{F}}_{\text{X}}}$ to $\mathbf{F}_{\text{HSI}}^{\text{r}}\in {{\mathbb{R}}^{{{s}^{2}}\times \tfrac{hw}{{{s}^{2}}}\times c}}$ and $\mathbf{F}_{\text{X}}^{\text{r}}\in {{\mathbb{R}}^{{{s}^{2}}\times \tfrac{hw}{{{s}^{2}}}\times c}}$, respectively. 

We first perform the region-to-region attention operation to select the most relevant regions. Since the operation of the X modality is identical to that of the HSI modality, we only present the process for the X modality for simplicity. For the X modality, the query, key, value tensors ${{\mathbf{Q}}_{\text{X}}},\ {{\mathbf{K}}_{\text{X}}},{{\mathbf{V}}_{\text{X}}}\in {{\mathbb{R}}^{{{s}^{2}}\times \tfrac{hw}{{{s}^{2}}}\times c}}$ are derived via ${{\mathbf{Q}}_{\text{X}}}=\mathbf{F}_{\text{X}}^{\text{r}}\mathbf{W}_{\text{X}}^{q}$, ${{\mathbf{K}}_{\text{X}}}=\mathbf{F}_{\text{X}}^{\text{r}}\mathbf{W}_{\text{X}}^{k}$ and ${{\mathbf{V}}_{\text{X}}}=\mathbf{F}_{\text{X}}^{\text{r}}\mathbf{W}_{\text{X}}^{v}$, where $\mathbf{W}_{\text{X}}^{q}$, $\mathbf{W}_{\text{X}}^{k}$ and $\mathbf{W}_{\text{X}}^{v}$ are the learnable projection weights for the query, key, value, respectively. Subsequently, we derive region-level queries and keys, $\mathbf{Q}_{\text{X}}^{\text{r}},\ \mathbf{K}_{\text{X}}^{\text{r}}\in {{\mathbb{R}}^{{{s}^{2}}\times c}}$, by averaging ${{\mathbf{Q}}_{\text{X}}}$ and ${{\mathbf{K}}_{\text{X}}}$ within each region. Then, the region-to-region adjacency matrix $\mathbf{A}_{\text{X}}^{\text{r}}\in {{\mathbb{R}}^{{{s}^{2}}\times {{s}^{2}}}}$ is obtained by:
\begin{equation}
\mathbf{A}_{\text{X}}^{\text{r}}=\mathbf{Q}_{\text{X}}^{\text{r}}{{\left( \mathbf{K}_{\text{X}}^{\text{r}} \right)}^{\text{T}}}.
\end{equation}
Each element in $\mathbf{A}_{\text{X}}^{\text{r}}$ measures the feature correlation between two regions. According to $\mathbf{A}_{\text{X}}^{\text{r}}$, the top-$k$ most relevant regions $\mathbf{I}_{\text{X}}^{\text{r}}\in {{\mathbb{R}}^{{{s}^{2}}\times k}}$ are selected as follows:
\begin{equation}
\mathbf{I}_{\text{X}}^{\text{r}}=\text{TopkIndex}\left( \mathbf{A}_{\text{X}}^{\text{r}} \right),
\end{equation}
where the $i$-th row of $\mathbf{I}_{\text{X}}^{\text{r}}$ contains the indices of the $i$-th most relevant regions.

Next, the fine-grained token-to-token attention is conducted, as shown in Fig.~\ref{FIFM}. For each query token in region $i$, it attends to all key-value pairs in the top-$k$ regions indexed with $\mathbf{I}_{\text{X}\left( i,1 \right)}^{r},\ \mathbf{I}_{\text{X}\left( i,2 \right)}^{r},\ \cdots ,\ \mathbf{I}_{\text{X}\left( i,k \right)}^{r}$. Since these top-$k$ regions are scattered over the entire feature map, the key and value tensors are first collected as follows:
\begin{equation}
\mathbf{K}_{\text{X}}^{g}=\text{Collect}\left( {{\mathbf{K}}_{\text{X}}},\mathbf{I}_{\text{X}}^{r} \right),
\end{equation}
\begin{equation}
\mathbf{V}_{\text{X}}^{g}=\text{Collect}\left( {{\mathbf{V}}_{\text{X}}},\mathbf{I}_{\text{X}}^{r} \right),
\end{equation}
where $\mathbf{K}_{\text{X}}^{g},\ \mathbf{V}_{\text{X}}^{g}\in {{\mathbb{R}}^{{{s}^{2}}\times \tfrac{khw}{{{s}^{2}}}\times c}}$ denote the collected key and value tensors of the top-$k$ regions for the X modality. 
For the HSI modality, the key and value tensors of the top-$k$ regions can also be obtained using the same operation as for the X modality, where they are represented as $\mathbf{K}_{\text{HSI}}^{g},\ \mathbf{V}_{\text{HSI}}^{g}\in {{\mathbb{R}}^{{{s}^{2}}\times \tfrac{khw}{{{s}^{2}}}\times c}}$, respectively. 

Cross-modality attention is then applied to the collected key-value pairs as follows:
\begin{equation}
{{\mathbf{O}}_{\text{HSI}}}=\text{Attention}\left( {{\mathbf{Q}}_{\text{HSI}}},\mathbf{K}_{\text{HSI}}^{g},\mathbf{V}_{\text{X}}^{g} \right),
\end{equation}

\begin{equation}
{{\mathbf{O}}_{\text{X}}}=\text{Attention}\left( {{\mathbf{Q}}_{\text{X}}},\mathbf{K}_{\text{X}}^{g},\mathbf{V}_{\text{HSI}}^{g} \right).
\end{equation}

The outputs of ${{\mathbf{O}}_{\text{HSI}}}$ and ${{\mathbf{O}}_{\text{X}}}$ are concatenated and fed into the DWFFN block to produce the fused features, i.e.
\begin{equation}
\mathbf{Y}=\text{DWFFN}\left[ {{\mathbf{O}}_{\text{HSI}}},\ {{\mathbf{O}}_{\text{X}}} \right].
\end{equation}
The output fused features $\mathbf{Y}$  are subsequently sent into the decoder for prediction.

\subsection{Decoder}
Enlarging the receptive field to capture more contextual information is a central concern in semantic segmentation. Over the past few years, most segmentation research has focused on designing decoders to enlarge the receptive field, such as the DeepLab series \cite{rDeepLab,rDeepLabV3} and PSPNet \cite{rPSPNet}. However, these approaches inevitably lead to heavier and more complex models. Considering that our encoder with transformers allows for a larger effective receptive field, we adopt the lightweight All-MLP decoder developed in \cite{rSegFormer}. 
As illustrated in Fig.~\ref{Overview}, the All-MLP decoder only consists of several MLP layers. The MLP layer is first applied to multi-level features from different stages to unify their channel dimension. The resulting feature maps are upsampled to match the resolution of the input image and then concatenated. The concatenated futures pass through two additional MLP layers for fusion and prediction, respectively. Notably, the decoder avoids computationally intensive components, such as atrous spatial pyramid pooling (ASPP) \cite{rDeepLab} or even $3\times 3$ convolution. This design choice allows our LoGoCAF to have a lower computational cost, fewer parameters, and increased efficiency.

\section{EXPERIMENTS AND RESULTS}
This section details the experimental datasets and settings, including compared methods, parameter settings, and evaluation metrics. Additionally, both quantitative and qualitative analyses of the experimental results are presented.

\begin{table}[th]
 \centering
  \caption{The land-cover types and the number of training and test samples in the Houston2013 dataset}
 \begin{tabular}{ccccc}
 \toprule[1pt]
    ID    & Color & Land-cover Type & Train & Test \\ \midrule[0.5pt]
    C1    & \cellcolor[rgb]{ 0,  .8,  0} & Healthy grass & 198   & 1053 \\
    C2    & \cellcolor[rgb]{ .498,  1,  0} & Stressed grass & 190   & 1064 \\
    C3    & \cellcolor[rgb]{ .18,  .541,  .341} & Artificial turf & 192   & 505 \\
    C4    & \cellcolor[rgb]{ 0,  .275,  0} & Evergreen trees & 188   & 1056 \\
    C5    & \cellcolor[rgb]{ .624,  .322,  .176} & Deciduous trees & 186   & 1056 \\
    C6    & \cellcolor[rgb]{ 0,  1,  1} & Bare earth & 182   & 143 \\
    C7    &                             & Water & 196   & 1072 \\
    C8    & \cellcolor[rgb]{ .851,  .745,  .922} & Residential buildings & 191   & 1053 \\
    C9    & \cellcolor[rgb]{ 1,  0,  0} & Non-residential buildings & 193   & 1059 \\
    C10   & \cellcolor[rgb]{ .314,  0,  0} & Roads & 191   & 1036 \\
    C11   & \cellcolor[rgb]{ 1,  .847,  .694} & Sidewalks & 181   & 1054 \\
    C12   & \cellcolor[rgb]{ 1,  1,  0} & Crosswalks & 192   & 1041 \\
    C13   & \cellcolor[rgb]{ .929,  .6,  0} & Major thoroughfares & 184   & 285 \\
    C14   & \cellcolor[rgb]{ .569,  .118,  .706} & Highways & 181   & 247 \\
    C15   & \cellcolor[rgb]{ 0,  .361,  1} & Railways & 187   & 473 \\   \midrule[0.5pt]
    \multicolumn{3}{c}{Total}  & 28332 & 12197 \\
\bottomrule[1pt]
\end{tabular}
\label{Houston_Tab}
\end{table}

\begin{figure}[h] 
\subfigure[]
{
\begin{minipage}[hb]{1\linewidth}
\flushleft
\includegraphics[width=8.5cm]{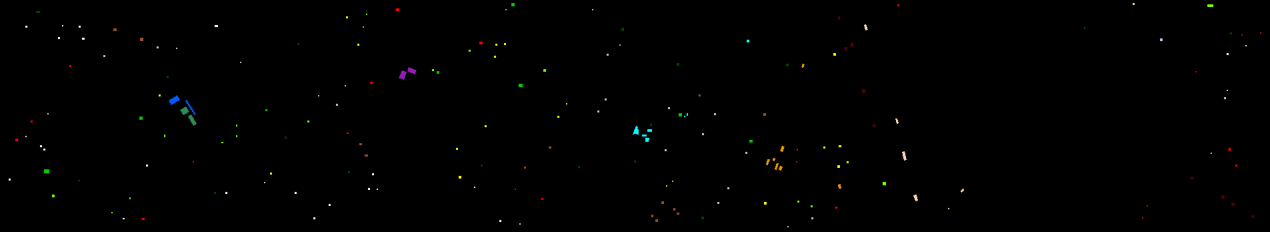}
\end{minipage}
}
\subfigure[]
{
\begin{minipage}[b]{1\linewidth}
\flushleft
\includegraphics[width=8.5cm]{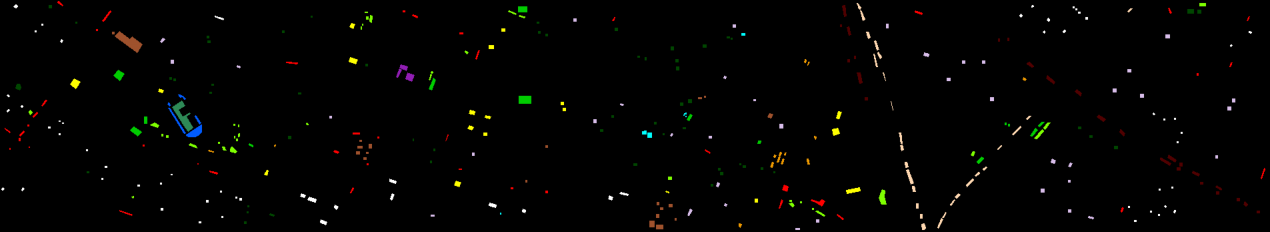}
\end{minipage}
}
\caption{Houston2013 dataset: spatial distribution of (a) training samples and (b) test samples.}
\label{Houston_Trn} 
\end{figure}

\subsection{Description of Datasets} 
We evaluate the effectiveness of our LoGoCAF framework through experiments conducted on three datasets that combines HSI-DSM, HSI-SAR, and HSI-multispectral LiDAR (MS-LiDAR) data.

The Houston2013 dataset, acquired by the National Centre for Airborne Laser Mapping in June 2012, covers the University of Houston campus and its adjacent urban areas. This dataset contains two data sources: an HSI and a LiDAR-derived DSM, both with the same spatial dimension of $349\times1905$ and spatial resolution of 2.5 m. The HSI has 144 spectral bands, covering the wavelength range from 380 nm to 1050 nm. Within this scene, 15 classes are defined. Table~\ref{Houston_Tab} summarizes the type and number of samples, while the spatial distribution of the training and test data is represented in Figs.~\ref{Houston_Trn}(a) and (b), respectively.

\begin{table}[th] 
 \centering
 \caption{The land-cover types and the number of training and test samples in the Berlin dataset}
 \begin{tabular}{ccccc}
 \toprule[1pt]
    ID    & Color & Land-cover Type & Train & Test \\ \midrule[0.5pt]
    C1    & \cellcolor[rgb]{ 0,  .4,  0} & Forest & 443   & 54511 \\
    C2    & \cellcolor[rgb]{ .89,  .89,  .89} & Residential Area & 423   & 268219 \\
    C3    & \cellcolor[rgb]{ .549,  .263,  .18} & Industrial Area & 499   & 19067 \\
    C4    & \cellcolor[rgb]{ .451,  1,  .675} & Low Plants & 376   & 58906 \\
    C5    & \cellcolor[rgb]{ 1,  1,  .49} & Soil  & 331   & 17095 \\
    C6    & \cellcolor[rgb]{ .235,  .353,  .447} & Allotment & 280   & 13025 \\
    C7    & \cellcolor[rgb]{ .733,  .082,  .906} & Commercial Area & 298   & 24526 \\
    C8    & \cellcolor[rgb]{ 0,  .361,  1} & Water & 170   & 6502 \\ \midrule[0.5pt]
    \multicolumn{3}{c}{Total}  & 2820 & 461851 \\
\bottomrule[1pt]
\end{tabular}
\label{Berlin_Tab}
\end{table}

\begin{figure}[h] 
\subfigure[]
{
\begin{minipage}[hb]{1\linewidth}
\centering
\includegraphics[width=8cm]{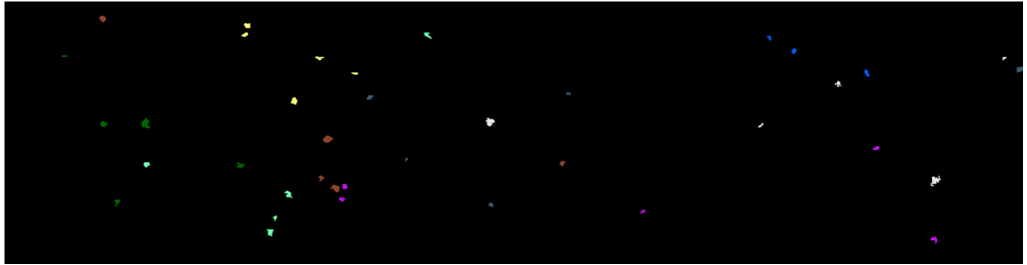}
\end{minipage}
}
\subfigure[]
{
\begin{minipage}[b]{1\linewidth}
\centering
\includegraphics[width=8cm]{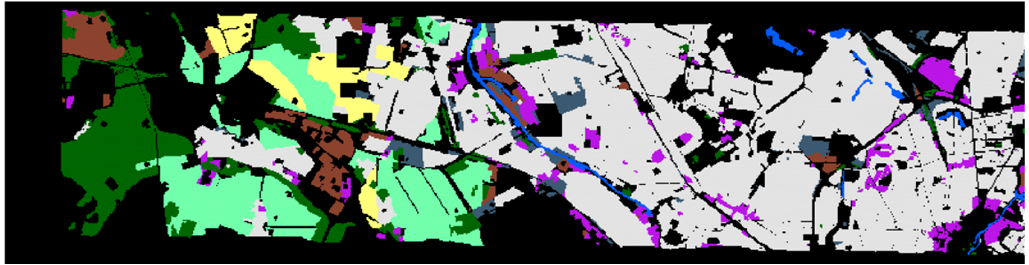}
\end{minipage}
}
\caption{Berlin dataset: spatial distribution of (a) training samples and (b) test samples.}
\label{Berlin_trn} 
\end{figure}

The Berlin dataset, acquired over the urban areas of Berlin and its surrounding rural areas, consists of HSI and SAR data. The HSI was generated using simulated EnMAP data synthesized from HyMap HSI data, while the SAR image was derived from the Sentinel 1 satellite dual-pol (VV-VH) single-look complex data. The HSI is a $797\times 220\times 224$ data cube with a spatial resolution of 30 m and a wavelength range of 0.4–2.5 µm. In contrast, the SAR image consists of $1723\times 476$ pixels with a spatial resolution of 13.89 m. Consequently, the HSI underwent the nearest neighbor interpolation to match the spatial resolution of the SAR image. Data preprocessing was performed, including orbital profiling, radiometric calibration, denoising, speckle reduction, and topographic correction. Detailed land-cover classes and the corresponding sample numbers are listed in Table~\ref{Berlin_Tab} and visualized in Fig.~\ref{Berlin_trn}.

\begin{table}[ht] 
 \centering
 \caption{The land-cover types and the number of training and test samples in the DFC2018 dataset}
 \begin{tabular}{ccccc}
 \toprule[1pt]
    ID    & Color & Land-cover Type & Train & Test \\ \midrule[0.5pt]   
    C1    & \cellcolor[rgb]{ 0,  .8,  0} & Healthy grass & 39196 & 20000 \\
    C2    & \cellcolor[rgb]{ .498,  1,  0} & Stressed grass & 130008 & 20000 \\
    C3    & \cellcolor[rgb]{ .18,  .541,  .341} & Artificial turf & 2736  & 20000 \\
    C4    & \cellcolor[rgb]{ 0,  .541,  0} & Evergreen trees & 54322 & 20000 \\
    C5    & \cellcolor[rgb]{ 0,  .275,  0} & Deciduous trees & 20172 & 20000 \\
    C 6   & \cellcolor[rgb]{ .624,  .322,  .176} & Bare earth & 18064 & 20000 \\
    C7    & \cellcolor[rgb]{ 0,  1,  1} & Water & 1064  & 1628 \\
    C8    &       & Residential buildings & 158995 & 20000 \\
    C9    & \cellcolor[rgb]{ .851,  .745,  .922} & Non-residential buildings & 894769 & 20000 \\
    C10   & \cellcolor[rgb]{ 1,  0,  0} & Roads & 183283 & 20000 \\
    C11   & \cellcolor[rgb]{ .663,  .624,  .584} & Sidewalks & 136035 & 20000 \\
    C12   & \cellcolor[rgb]{ .498,  .498,  .498} & Crosswalks & 6059  & 5345 \\
    C13   & \cellcolor[rgb]{ .624,  0,  0} & Major thoroughfares & 185438 & 20000 \\
    C14   & \cellcolor[rgb]{ .314,  0,  0} & Highways & 39438 & 20000 \\
    C15   & \cellcolor[rgb]{ 1,  .847,  .694} & Railways & 27748 & 11232 \\
    C16   & \cellcolor[rgb]{ 1,  1,  0} & Paved parking lots & 45932 & 20000 \\
    C17   & \cellcolor[rgb]{ .929,  .6,  0} & Unpaved parking lots & 587   & 3524 \\
    C18   & \cellcolor[rgb]{ 1,  0,  1} & Cars  & 26289 & 20000 \\
    C19   & \cellcolor[rgb]{ 0,  0,  1} & Trains & 21479 & 20000 \\
    C20   & \cellcolor[rgb]{ .686,  .765,  .867} & Stadium seats & 27296 & 20000 \\  \midrule[0.5pt]
    \multicolumn{3}{c}{Total}  & 2018910 & 341729 \\
\bottomrule[1pt]
\end{tabular}
\label{DFC_Tab}
\end{table}

\begin{figure}[hbt] 
{
\begin{minipage}[hb]{1\linewidth}
\centering
\includegraphics[width=8.5cm]{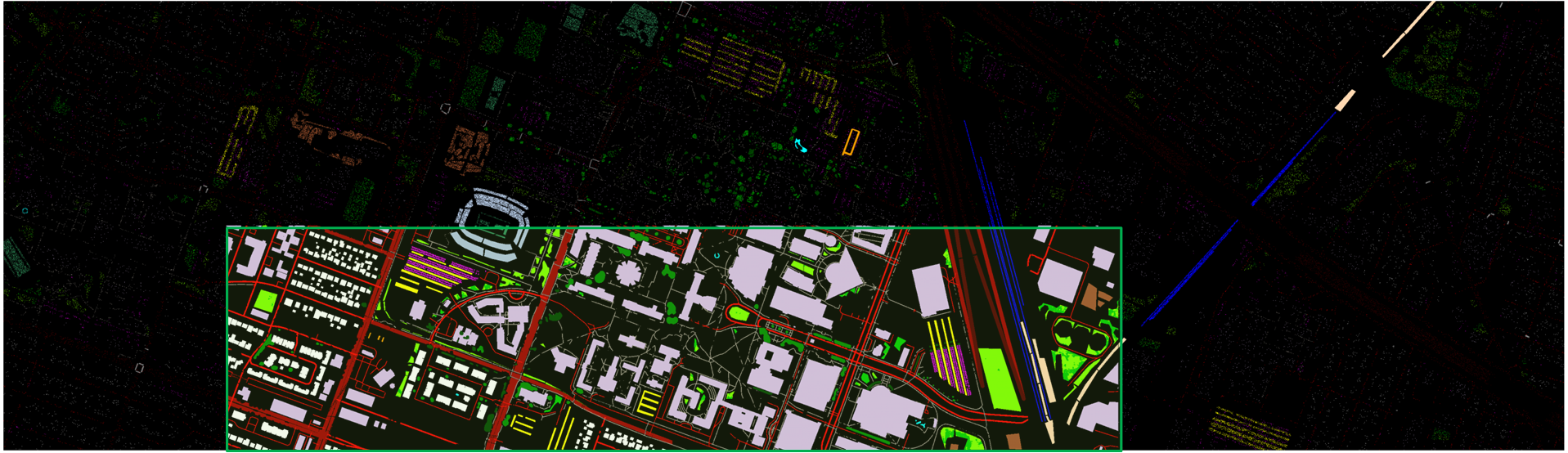}
\end{minipage}
}
\caption{DFC2018 dataset: spatial distribution of training (green box) and test samples (outside the green box).}
\label{DFC_trn} 
\end{figure}

The DFC2018 dataset, acquired by the National Center for Airborne Laser Mapping in February 2017, covers the University of Houston campus and its surrounding urban areas. The data used in this study consists of HSI data with a 1-m ground sampling distance (GSD) and MS-LiDAR point cloud data with a 0.5-m GSD. The HSI consists of $4172\times 1202$ pixels across 48 bands, spanning the spectral range from 380 nm to 1050 nm. Conversely, the MS-LiDAR data includes three intensity rasters at different wavelengths: 1550 nm (near infrared), 1064 nm (near infrared), and 532 nm (green), along with a digital surface model (DSM) image and a digital elevation model (DEM) image. To match with the spatial resolution of the MS-LiDAR images, the HSI  were resampled to a 0.5-m GSD using the nearest neighbor approach. We calculate the Normalized DSM (NDSM) value via $\mathrm{NDSM}=\mathrm{DSM}-\mathrm{DEM}$ to obtain the actual elevation of the objects. As summarized in Table~\ref{DFC_Tab}, this dataset includes 20 land-cover classes. The spatial distribution of the training and test data is shown in Fig.~\ref{DFC_trn}.

Prior to the experiments, we normalized these datasets to [0, 1] to standardize the data magnitude and thereby facilitate network convergence.

\subsection{Experimental Settings} 

\begin{table*}[tbp]
  \centering
  \caption{Comparison among the classification accuracy of different methods on the Houston2013 dataset. The numbers within ( ) are the standard deviations of the corresponding metrics. The best result in each row is marked in bold.}
   \begin{tabular}{p{4.19em}cccccccccccc}
    \toprule[1pt]
    \makecell{Metrics \\ and Class} & \makecell{SVM-X \\ \cite{r4}}  & \makecell{SVM-HSI \\ \cite{r4}} & \makecell{SVM \\ \cite{r4}} & \makecell{S2FL \\ \cite{8718389}} & \makecell{FusAtNet \\ \cite{9150738}} & \makecell{CACL \\ \cite{rCALC}} & \makecell{Fusion\_HCT \\ \cite{rFusionHCT}} & \makecell{MFT \\ \cite{rMFT}} & \makecell{Fusion-FCN \\ \cite{rFusionFCN}}  & \makecell{LoGoCAF \\ (Ours)} \\
    \midrule[0.5pt] 

     OA (\%)  & \makecell{17.16  \\ (0)} & \makecell{55.51 \\ (0)} & \makecell{64.68 \\ (0)} & \makecell{82.40 \\ (0)} & \makecell{89.02\\ (1.03)} & \makecell{86.24 \\ (1.56)} & \makecell{88.95 \\ (1.48)} & \makecell{87.76 \\ (0.98)}  & \makecell{83.84 \\ (1.67)}  & \makecell{\textbf{92.11} \\ (1.18)} \\
    AA (\%)  & \makecell{16.66 \\ (0)} & \makecell{59.83 \\ (0)} & \makecell{67.14 \\ (0)} & \makecell{84.73\\ (0)} & \makecell{92.28\\ (1.18)} & \makecell{88.10 \\ (2.18)} & \makecell{90.48 \\ (1.91)} & \makecell{89.28 \\ (1.06)} &  \makecell{85.42 \\ (2.16)}  & \makecell{\textbf{93.33} \\ (1.06)} \\
    $ \kappa\times 100$ & \makecell{10.76 \\ (0)} & \makecell{52.12 \\ (0)} & \makecell{61.82 \\ (0)} & \makecell{80.88 \\ (0)} & \makecell{88.09 \\ (1.14)} & \makecell{85.11 \\ (1.92)} & \makecell{88.00 \\ (1.62)} & \makecell{86.42 \\ (0.98)} & \makecell{82.50 \\ (1.61)}  & \makecell{\textbf{91.44} \\ (1.36)} \\
    \midrule[0.5pt]

    C1 & 55.56  & 81.96  & 82.43  & 81.77  & 82.96 & 80.63  & 82.48  & 82.67  & 83.00  & \textbf{83.10} \\
    C2 & 0.00  & 75.00  & 80.83  & 83.55  & \textbf{97.70} & 80.83  & 83.93  & 84.48  & 84.02  & 85.06  \\
    C3 & \textbf{100.00} & 99.01  & 99.01  & \textbf{100.00} & \textbf{100.00} & 85.55  & 98.22  & 98.68  & \textbf{100.00} & \textbf{100.00} \\
    C4 & 15.63  & 89.21  & 88.35  & 94.51  & 95.74  & 91.00  & 92.36  & 92.33 & 90.72  & \textbf{100.00} \\
    C5 & 0.00  & 85.23  & 90.06  & 99.43  & 98.96  & 99.34  & 99.91  & 98.22  & 99.34  & \textbf{100.00} \\
    C6 & 0.00  & 78.32  & 78.32  & 99.30  & \textbf{100.00} & 93.01  & 91.72  & 95.45  & 99.30  & \textbf{100.00} \\
    C7 & 47.30  & 28.64  & 69.50  & 76.68  & \textbf{92.19} & 83.12  & 87.35  & 85.11  & 82.46  & 86.29  \\
    C8 & 31.43  & 12.92  & 54.42  & 75.21  & 82.81 & 85.28  & \textbf{91.19} & 78.73  & 90.98  & 83.95  \\
    C9 & 0.00  & 81.30  & 83.00  & 72.24  & 85.35 & 86.03  & 79.55  & \textbf{87.73} & 75.54  & 80.55  \\
    C10 & 0.00  & 1.83  & 2.90  & 49.13  & 66.12  & 60.91  & 67.31  & 60.94  & 54.83  & \textbf{94.40} \\
    C11 & 0.00  & 55.88  & 64.52  & 89.47  & 85.06 & 94.88  & 96.40  & 96.01 & 92.60  & \textbf{99.24} \\
    C12 & 0.00  & 0.10  & 0.00  & 83.38  & 89.63  & 88.38  & 95.89  & 92.88  & 70.70  & \textbf{100.00} \\
    C13 & 0.00  & 11.93  & 15.09  & 67.72  & 86.32  & \textbf{92.98} & 90.88  & 90.01  & 59.30  & 87.72  \\
    C14 & 0.00  & 97.17  & 99.60  & \textbf{100.00}  & 99.19  & 99.60  & \textbf{100.00} & 99.87 & \textbf{100.00} & \textbf{100.00} \\
    C15   & 0.00  & 98.94  & 99.15  & 98.52  & 99.58 & \textbf{100.00} & \textbf{100.00} & 96.09  & 98.52  & 99.58  \\
    \bottomrule[1pt]
    \end{tabular}%
  \label{Houston_acc}
\end{table*}%

\subsubsection{compared methods}
We evaluated our LoGoCAF framework against traditional techniques: SVM \cite{r4} and S2FL \cite{8718389}. Additionally, we compared it with several deep learning-based models: FusAtNet \cite{9150738}, coupled adversarial learning-based classification (CALC) \cite{rCALC}, Fusion\_HCT \cite{rFusionHCT}, MFT \cite{rMFT}, and Fusion-FCN \cite{rFusionFCN}. SVM is a traditional pixel-wise classifier, where data are combined by pixel-level fusion. S2FL aligns the shared components between multimodalities on manifolds, decoupling different modalities into shared and specific feature spaces. FusAtNet, CALC, Fusion\_HCT, and MFT are designed for fusing HSI and LiDAR data within a patch-based classification framework. Specifically, FusAtNet generates an HSI-derived attention map and a LiDAR-derived attention map that highlight the spectral and spatial features of HSIs, respectively. CACL uses a coupled adversarial feature learning subnetwork to learn high-order semantic features from HSI and LiDAR data, and then exploits a multi-level feature fusion classification subnetwork for decision fusion and classification. Fusion\_HCT sequentially performs feature extraction with CNN and transformer blocks, followed by a cross-token attention fusion structure for feature fusion. Meanwhile, MFT adopts a ViT-based network, in which the input patches are taken from HSIs, while the class token is generated from the LiDAR data. Fusion-FCN is an FCN-based architecture in which the feature maps of each layer share a consistent spatial size with the input images. Note that since other compared methods did not use any post-processing, for a fair comparison, we only compared with the first-place winner of the 2018 IEEE DFC without post-classification processing, i.e., Fusion-FCN.

\subsubsection{Implementation Details}
In LoGoCAF, there are four stages in total. Typically, a downsampling layer is employed after each stage to capture multi-scale features while reducing the computational load. However, experimental results reveal that the highest accuracy is achieved when the downsampling layer is applied only after the first stage. This observation suggests that increasing the number of downsampling layers reduces the computational load, but it also results in the loss of significant spatial information, especially for small and fine objects, such as roads, sidewalks, and stadium seats in the DFC2018 datasets. This phenomenon can be attributed to the trade-off between computational efficiency and spatial information extraction. Therefore, in LoGoCAF, we only perform 2× downsampling between the first and second stages.

We trained our LoGoCAF network using the Adam optimizer and the cross-entropy loss function with a weight decay of 0.01 and an initial learning rate of $6\times 10^{-5}$. The learning rate is updated during training using the poly-learning rate schedule. Considering the limited memory of the GPU used in our experiments, we partitioned the entire image into $128\times 128$ sub-images with overlay ratio of 50$\%$ to reduce memory cost. The batch size and the number of training epochs were set to 4 and 500, respectively. We will save the best performing model from the training process to use for testing and prediction. 

For the compared methods, we used their publicly available source codes and set their parameters according to the relevant literature. All compared methods were implemented on the PyTorch platform and trained and tested on the same sample sets, as listed in Tables.~\ref{Houston_Tab}–\ref{DFC_Tab}. To ensure unbiased results, we performed five independent tests for each experiment. The average evaluation metrics are reported to provide a comprehensive and accurate insight into our findings.

\subsubsection{Metric}
We use the overall accuracy (OA), average accuracy (AA), kappa coefficient ($\kappa $), and producer accuracy (PA) of each category as the primary evaluation metrics to assess the performance of different approaches. 

\subsection{Comparison with Other Methods} 
\subsubsection{Quantitative Results and Analysis}

\begin{figure*}[tb]      
\centerline{\includegraphics[width=17.5cm]{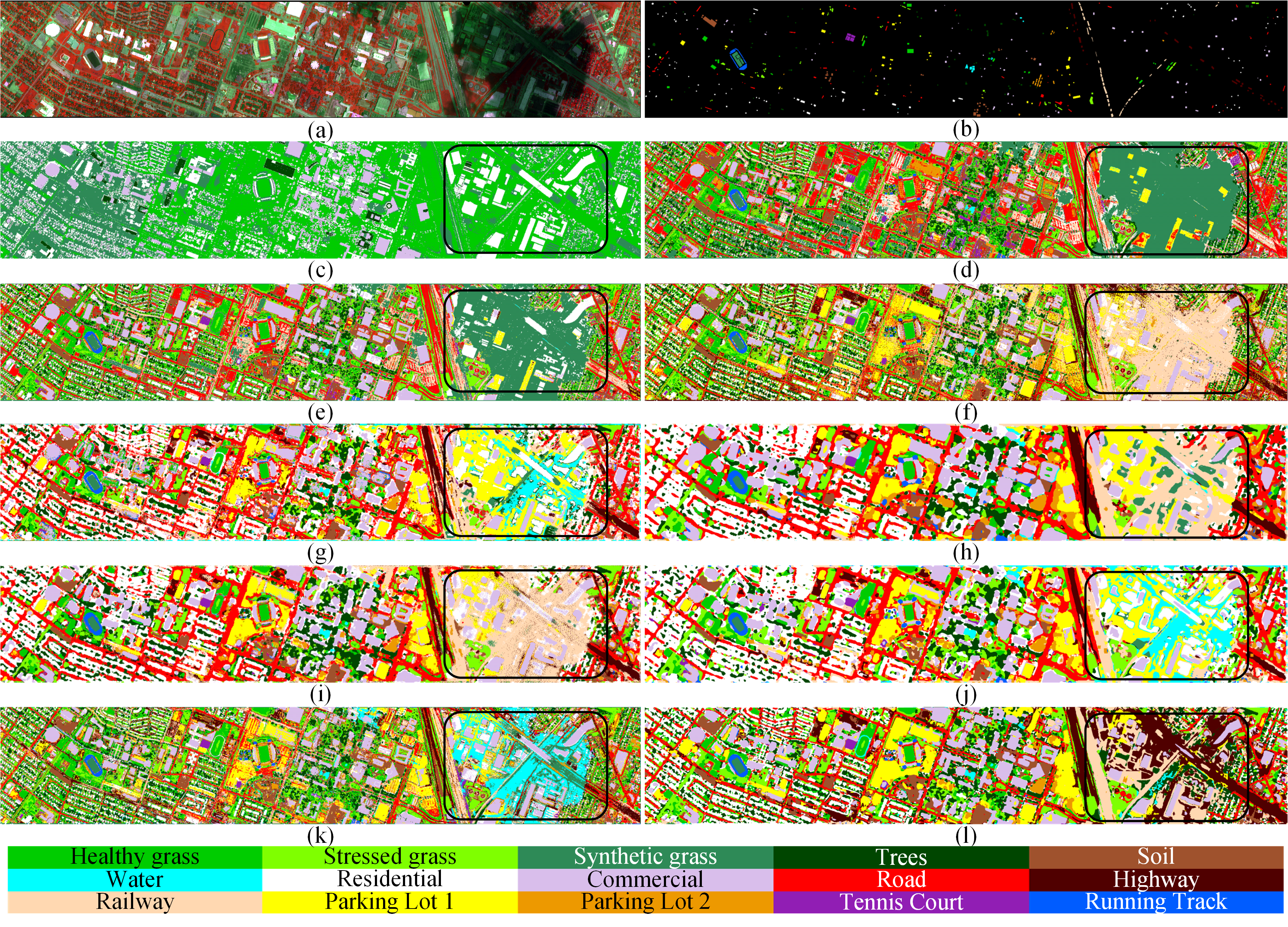}} 
\caption{Classification maps of different methods on the Houston2013 dataset: (a) False Color Image, (b) Ground Truth, (c) SVM-X, (d) SVM-HSI, (e) SVM, (f) S2FL, (g) FusAtNet, (h) CACL, (i) Fusion\_HCT, (j) MFT, (k) Fusion-FCN and (l) LOGoCAF.}
\label{Houston_map} 
\end{figure*}

The mean and standard deviation of OA, AA, and $ \kappa $, along with the mean PA of each category produced by the compared methods are summarized in Tables~\ref{Houston_acc}–\ref{DFC_acc}, with the best result in each row is marked in bold. In Tables~\ref{Houston_acc}–\ref{DFC_acc}, SVM-X, SVM-HSI, and SVM denote that SVM classifies HSI, X, and HSI-X data, respectively. Conversely, the other methods all focus on classifying HSI-X data together. Across all three datasets, our LoGoCAF consistently outperforms the others in terms of OA and Kappa. Taking Table~\ref{Houston_acc} as an example, LoGoCAF achieves the highest OA of 92.11$\%$, surpassing SVM, S2FL, FusAtNet, CALC, Fusion\_HCT, MFT, and Fusion-FCN by 27.43$\%$, 9.71$\%$, 3.09$\%$, 5.87$\%$, 3.16$\%$, 4.35$\%$, and 8.27$\%$, respectively. 

As indicated in Tables~\ref{Houston_acc}–\ref{DFC_acc}, all compared methods achieve the best OA on the Houston2013 dataset, generally over 80$\%$, followed by the Berlin dataset, while the accuracy on the DFC2018 dataset is the lowest at around 60$\%$. This phenomenon is particularly evident with FusAtNet, CACL and Fusion\_HCT. This discrepancy in accuracy can be attributed to two primary factors. Firstly, the land-cover categories defined in the DFC2018 dataset are more refined, making it challenging to distinguish certain objects with similar or even identical material, such as road, major thoroughfares and highways. Secondly, the distributions of training and test samples differ between datasets. The test samples are scattered around the training samples on the Houston2013 and Berlin datasets, while the test and training samples are completely separated on the DFC2018 dataset. It will result in overlapping feature spaces between training and test samples when the distance between them is close. Thus, the higher accuracy observed on the Houston2013 and Berlin datasets is partly due to the bias introduced by the increased dependency between training and test samples, highlighting the impact of sample distribution on performance evaluation. Furthermore, the performance of SVM on HSI data significantly outperforms its performance on X data by a remarkable margin of 38.35$\%$ on Houston2013, 33.80$\%$ on Berlin, and 15.48$\%$ on DFC2018 data. This confirms that, as expected, the HSI data contain more discriminative features than the X data. Combining HSI and X data, the accuracy of SVM on the Houston2013, Berlin and DFC2018 datasets is further improved to 64.68$\%$, 62.98$\%$, and 56.60$\%$, respectively. This indicates that elevation information, SAR and MS-LiDAR play a significant role in the classification task.

Although SVM achieves better performance after combining HSI and X data, it still lags behind other compared methods, especially the DL-based methods. As a conventional pixel-level classifier, SVM directly processes raw pixels, lacking the ability to exploit discriminative information within the data. With the introduction of spatial information, S2FL significantly outperforms SVM by 17.72$\%$ in OA on the Houston2013 dataset, but slightly underperforms SVM by 0.75$\%$ in OA on the Berlin dataset. However, S2FL cannot produce results on the large dataset of DFC2018 due to the need to calculate pairwise distances between all training pixels, resulting in insufficient memory. 

\begin{table*}[htbp]
  \centering
  \caption{Comparison among the classification accuracy of different methods on the Berlin dataset. The numbers within ( ) are the standard deviations of the corresponding metrics. The best result in each row is marked in bold.}
   \begin{tabular}{p{4.19em}cccccccccccc}
    \toprule[1pt]
    
    \makecell{Metrics \\ and Class} & \makecell{SVM-X \\ \cite{r4}}  & \makecell{SVM-HSI \\ \cite{r4}} & \makecell{SVM \\ \cite{r4}} & \makecell{S2FL \\ \cite{8718389}} & \makecell{FusAtNet \\ \cite{9150738}} & \makecell{CACL \\ \cite{rCALC}} & \makecell{Fusion\_HCT \\ \cite{rFusionHCT}} & \makecell{MFT \\ \cite{rMFT}} & \makecell{Fusion-FCN \\ \cite{rFusionFCN}}  & \makecell{LoGoCAF \\ (Ours)} \\
    \midrule[0.5pt] 
     OA (\%)  & \makecell{27.54  \\ (0)} & \makecell{61.34 \\ (0)} & \makecell{62.98 \\ (0)} & \makecell{62.23\\ (0)} & \makecell{73.86\\ (1.84)} & \makecell{75.39 \\ (1.76)} & \makecell{76.46 \\ (1.51)} & \makecell{75.21 \\ (1.59)}  & \makecell{55.14 \\ (1.74)}  & \makecell{\textbf{77.33} \\ (1.32)} \\
    AA (\%)  & \makecell{20.11 \\ (0)} & \makecell{61.15 \\ (0)} & \makecell{62.37 \\ (0)} & \makecell{62.48\\ (0)} & \makecell{\textbf{66.21}\\ (2.08)} & \makecell{63.83 \\ (2.31)} & \makecell{61.47 \\ (2.13)} & \makecell{64.02\\ (2.19)} & \makecell{57.52 \\ (2.32)} & \makecell{64.69 \\ (0.98)} \\
    $ \kappa\times 100$ & \makecell{10.25 \\ (0)} & \makecell{47.59 \\ (0)} & \makecell{49.53 \\ (0)} & \makecell{48.77 \\ (0)} & \makecell{60.72 \\ (1.97)} & \makecell{61.89 \\ (2.08)} & \makecell{63.38 \\ (1.92)} & \makecell{63.16 \\ (1.61)} & \makecell{40.23 \\ (1.74)}  & \makecell{\textbf{64.16} \\ (1.27)} \\
    \midrule[0.5pt]
    C1    & \textbf{88.27} & 80.34  & 80.23  & 83.30  & 50.05  & 47.33  & 59.50  & 79.69  & 43.19  & 73.02  \\
    C2    & 26.18  & 57.41  & 58.34  & 57.39  & 81.09  & 86.43  & 85.89  & 79.19  & 54.44  & \textbf{88.75} \\
    C3    & 46.43  & 46.43  & 48.30  & 48.53  & 48.51  & 41.13  & \textbf{65.57} & 42.84  & 34.15  & 52.64  \\
    C4    & 0.00  & 76.36  & 84.88  & 77.16  & 85.58 & 80.30  & 85.57  & \textbf{87.50}  & 69.40  & 68.45  \\
    C5    & 0.00  & 74.24  & 74.40  & 83.84  & \textbf{96.38} & 96.27  & 83.54  & 70.64  & 78.15  & 80.95  \\
    C6    & 0.00  & 60.85  & 59.75  & 57.05  & 59.74  & 60.14  & 55.03  & 62.27  & \textbf{64.16} & 61.65  \\
    C7    & 0.00  & 27.84  & 27.25  & 31.02  & 29.88  & 26.09  & 12.72  & 32.06 & \textbf{46.77}  & 20.46  \\
    C8    & 0.00  & 65.73  & 65.80  & 61.57  & \textbf{78.42} & 72.95  & 43.92  & 57.98  & 69.92  & 71.58  \\
    \bottomrule[1pt]
    \end{tabular}%
  \label{Berlin_acc}
\end{table*}%

\begin{figure*}[th] 
\centerline{\includegraphics[width=16.5cm]{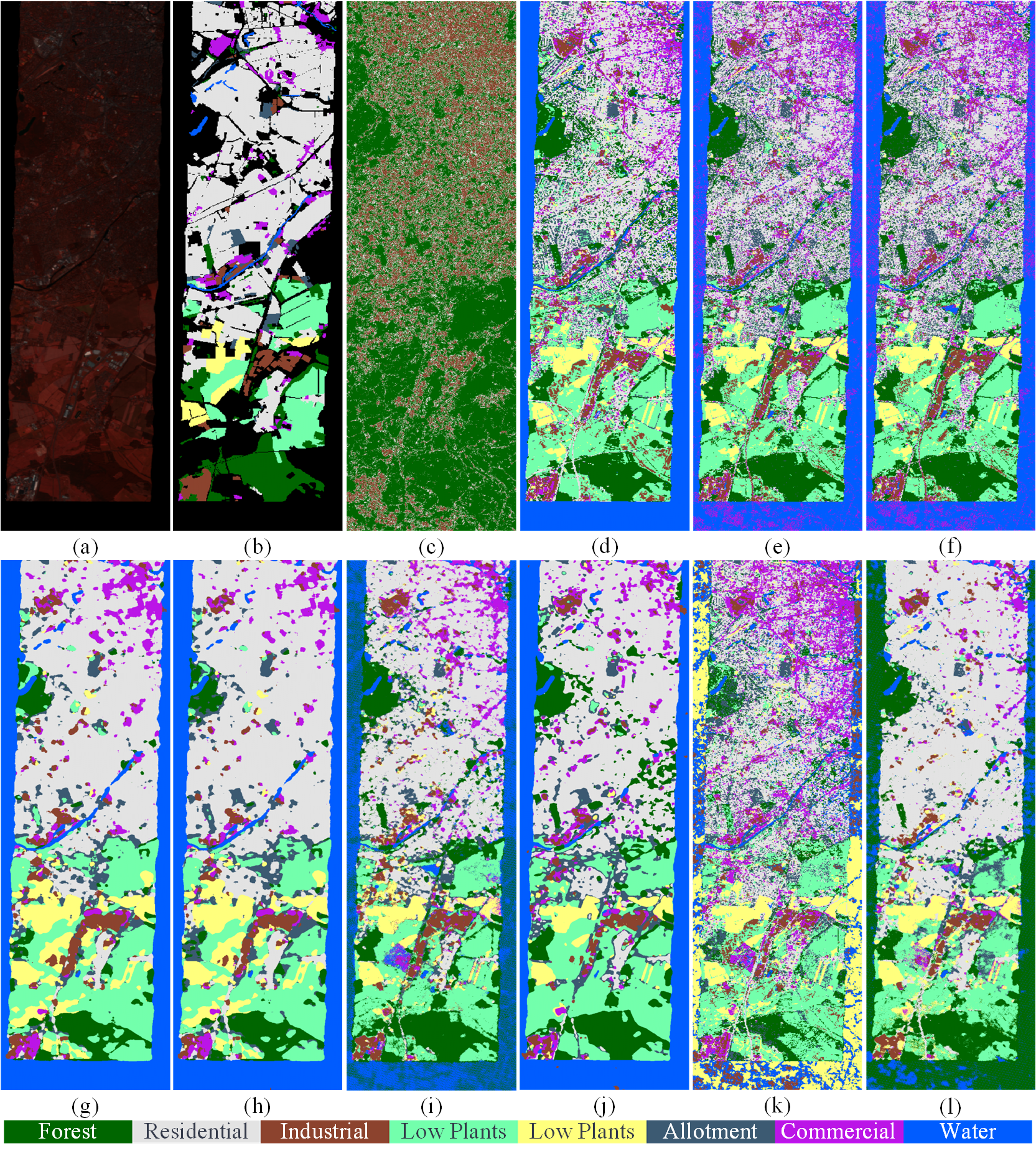}} 
\caption{Classification maps of different methods on the Berlin dataset:  (a) False Color Image, (b) Ground Truth, (c) SVM-X, (d) SVM-HSI, (e) SVM, (f) S2FL, (g) FusAtNet, (h) CACL, (i) Fusion\_HCT, (j) MFT, (k) Fusion-FCN and (l) LoGoCAF.}
\label{Berlin_map} 
\end{figure*}

In contrast, DL-based methods (i.e., FusAtNet, CACL, Fusion\_HCT, MFT, and Fusion-FCN) exhibit strong learning capabilities and consistently improve performance on all three datasets. For instance, these DL-based methods except Fusion-FCN all achieve over 80$\%$ and 73$\%$ OA on the Houston2013 and Berlin datasets, respectively. Although FusAtNet, CACL, Fusion\_HCT, and MFT achieve higher accuracy by using attention mechanisms to refine feature representations or to perform cross-modality fusion, their limited patch size prevents them from modeling long-range spatial information, resulting in the loss of some important information.

Overcoming the constraint of limited patch size, Fusion-FCN enables the processing of sufficiently large input images and performs pixel-to-pixel classification. However, its performance is on par with or even worse than those of patch-based networks. For instance, on the Berlin dataset, FusAtNet and CACL achieve OA of 73.86$\%$ and 75.39$\%$, respectively, while Fusion-FCN achieves only 55.14$\%$. We argue that this is because the Fusion-FCN only uses a stack of multiple 3 × 3 convolutions to enlarge the receptive fields, but its effective receptive field (ERF) remains limited according to the ERF theory \cite{rERF}, failing to learn adequate spatial features. 

Compared with the aforementioned approaches, our LoGoCAF exhibits significant advancements. Even on the challenging dataset of DFC2018, LoGoCAF still achieves superior accuracy by exploiting both local and long-range discriminative information across spectral and spatial dimensions. LoGoCAF not only adheres the pixel-to-pixel segmentation framework but also utilizes convolution to gather local information and transformers to model long-range dependencies. Moreover, our novel cross-modality FEM and FIFM facilitate the integration and utilization of cross-modality information.

\subsubsection{Qualitative  Results and Analysis}

\begin{table*}[htbp]
  \centering
  \caption{Comparison among the classification accuracy of different methods on the DFC2018 dataset. The numbers within ( ) are the standard deviations of the corresponding metrics. The best result in each row is marked in bold.}
   \begin{tabular}{p{4.19em}cccccccccccc}
    \toprule[1pt]
    \makecell{Metrics \\ and Class} & \makecell{SVM-X \\ \cite{r4}}  & \makecell{SVM-HSI \\ \cite{r4}} & \makecell{SVM \\ \cite{r4}} &  \makecell{FusAtNet \\ \cite{9150738}} & \makecell{CACL \\ \cite{rCALC}} & \makecell{Fusion\_HCT \\ \cite{rFusionHCT}} & \makecell{MFT \\ \cite{rMFT}} & \makecell{Fusion-FCN \\ \cite{rFusionFCN}}  & \makecell{LoGoCAF \\ (Ours)} & \makecell{LoGoCAF\\ (Ours on 1-m GSD)} \\
    \midrule[0.5pt] 
    OA (\%)  & \makecell{25.67  \\ (0)} & \makecell{41.15 \\ (0)} & \makecell{56.60 \\ (0)} & \makecell{61.24\\ 0)} & \makecell{60.06 \\ (1.56)} & \makecell{62.90\\ (1.18)} & \makecell{60.26 \\ (1.38)}  & \makecell{62.87 \\ (1.47)} & \makecell{\textbf{65.50} \\ (1.17)}  & \makecell{62.93 \\ (1.59)} \\
    AA (\%)  & \makecell{21.93 \\ (0)} & \makecell{36.45 \\ (0)} & \makecell{49.90 \\ (0)} & \makecell{56.17\\ (0)} & \makecell{53.26\\ (1.89)} & \makecell{56.84 \\ (2.11)} & \makecell{54.17 \\ (1.79)} & \makecell{57.00\\ (2.39)} & \makecell{\textbf{61.23} \\ (1.08)} & \makecell{55.83 \\ (2.12)} \\
    $ \kappa\times 100$ & \makecell{21.05 \\ (0)} & \makecell{37.50\\ (0)} & \makecell{53.91 \\ (0)}& \makecell{58.79 \\ (0)}  & \makecell{57.61\\ (1.50)} & \makecell{60.66 \\ (1.71} & \makecell{57.85 \\ (1.82)} & \makecell{60.62 \\ (1.71)} & \makecell{\textbf{63.47} \\ (1.26)} & \makecell{60.64\\ (1.21)} \\
    \midrule[0.5pt]
    C1    & 8.11  & 95.82  & 96.18  & 95.27  & 94.60  & 96.19  & 95.47  & 96.53  & \textbf{98.34}  & 94.37 \\
    C2    & 90.89  & 91.86 & 91.59  &87.13  & 89.99 & 86.01  & 90.86 & 87.80  &  84.13 &  \textbf{91.98} \\
    C3    & 0.00  & 40.50  & 54.81  & 15.72  & \textbf{60.15}  & 53.30  & 34.23  & 48.79  & 14.08 &  17.37  \\
    C4    & 66.37  & 86.95  & 95.71  & 97.28  & 94.43  & 95.39  & 95.50  & 95.99  &  \textbf{98.23} & 95.50 \\
    C5    & 0.00  & 28.02  & 55.32  & 57.97  & 48.75  & 53.00  & 49.70  & \textbf{60.02}  & 58.87  & 45.08 \\
    C6    & 0.00  & 26.72  & 49.12  & 85.20  & 45.86  & 56.58  & 53.93  & 62.17  & \textbf{88.34}  & 78.75 \\
    C7    & 0.00  & 28.13  & 33.35  & 81.23  & 31.70  & 56.05  & 49.55  & 59.37 & \textbf{99.82}  & 34.89 \\
    C8    & 59.82  & 24.34  & 35.32  & \textbf{79.32 } & 51.96  & 64.65  & 61.78  & 59.33  & 76.51 &  64.70  \\
    C9    & 87.02  & 85.53  & 88.34  & 92.15  & 90.72  & 91.04  & 90.93  & 89.46  &  91.95  & \textbf{93.64} \\
    C10   & 84.09  & 61.28  & 73.35  & 87.95  & 65.07  & 56.54  & 81.44  & 78.37  & \textbf{89.04} & 85.99   \\
    C11   & 39.57  & 35.86  & 53.98  & 64.73  & 56.84  & 59.94  & 63.31  & 60.62  & 66.73 &  \textbf{74.16} \\
    C12   & 0.00  & 0.00  & 0.00  & 8.61  & 9.36  & 9.02  & 7.11  & 10.05  &  \textbf{11.73} & 8.73 \\
    C13   & 2.53  & 44.05  & \textbf{49.84}  & 33.41  & 49.47  & 44.46  & 40.63  & 45.72  & 31.12  & 36.79  \\
    C14   & 0.00  & 8.52  & 9.45  & 14.22  & 24.81  & \textbf{26.75} & 20.67  & 21.65  & 18.92   & 18.95  \\
    C15   & 0.00  & 0.05  & 0.47  & 9.43  & 6.94  & 8.76  & 7.28  & 9.04  &  \textbf{11.39} & 7.48 \\
    C16   & 0.00  & 15.56  & 46.19  & \textbf{85.34} & 49.72  & 68.57  & 65.05  & 55.61  & 85.13  &  85.00  \\
    C17   & 0.00  & 0.00  & 0.00  & 0.00  & 0.00  & 0.00  & 0.00  & 0.00  & 0.00  & 0.00  \\
    C18   & 0.03  & 0.83  & 38.39  & 33.78  & 50.35  & 48.36  & 46.16  & \textbf{52.53}  &  46.44  & 37.97 \\
    C19   &  0.16  & 6.74  & 59.47  & 74.12  & 75.12  & 88.12  & 62.39  & 82.19  &  \textbf{90.47}  &  77.51\\
    C20   &  0.01  & 48.22  & 67.13  & 20.48  & 69.41  & \textbf{74.00} & 67.47  & 64.77  & 63.33  & 67.68  \\
    \bottomrule[1pt]
    \end{tabular}%
  \label{DFC_acc}
\end{table*}%

\begin{figure*}[htb] 
\centerline{\includegraphics[width=16.5cm]{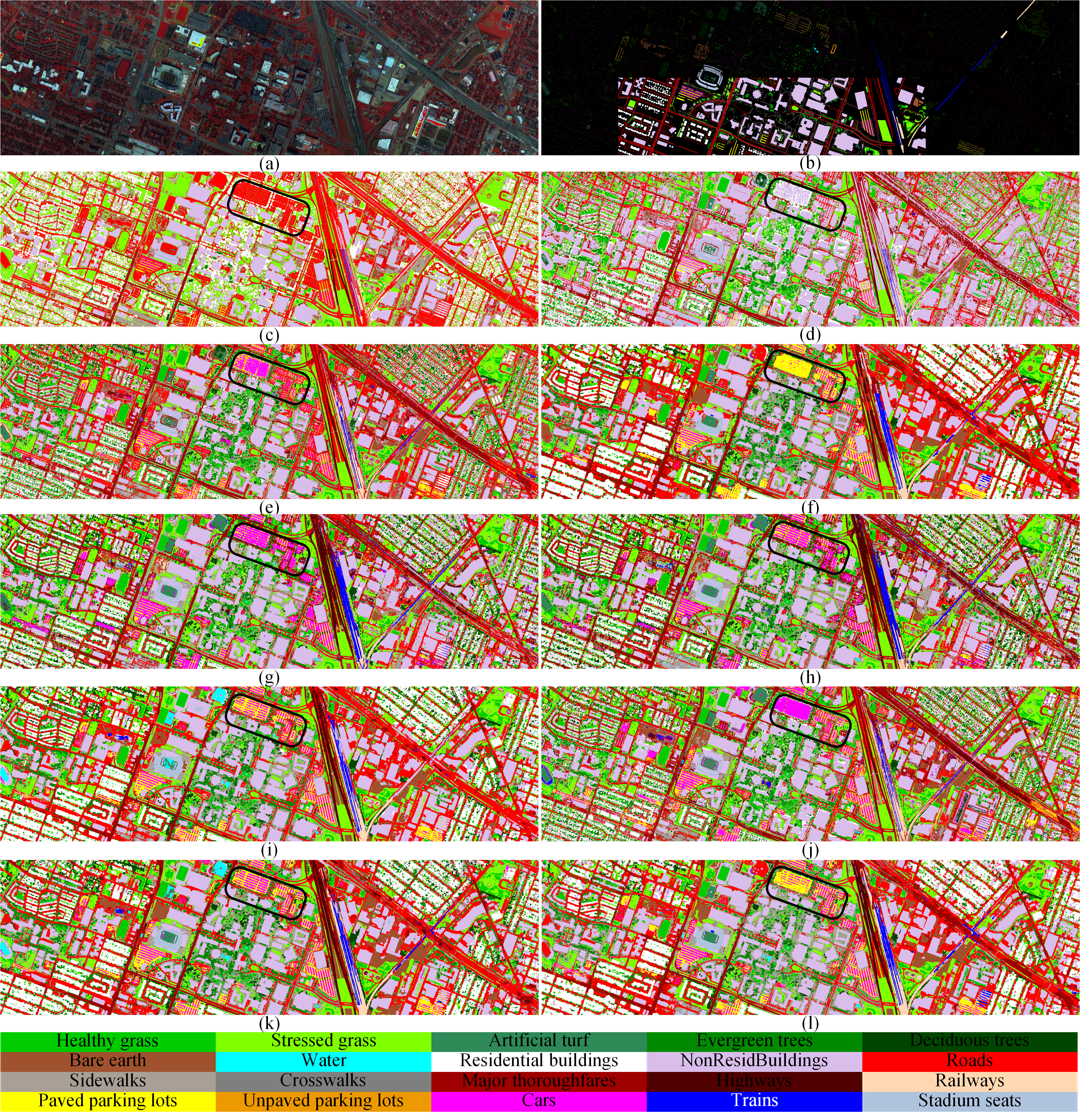}} 
\caption{Classification maps of different methods on the DFC2018 dataset:  (a) False Color Image, (b) Ground Truth, (c) SVM-X, (d) SVM-HSI, (e) SVM, (f) FusAtNet, (g) CACL, (h) Fusion\_HCT, (i) MFT, (j) Fusion-FCN, (k)  LoGoCAF and (l) LoGoCAF on 1-m GSD.}
\label{DFC_map}
\end{figure*}

The classification maps, along with the corresponding false-color and ground truth (GT) images, are presented in Figs.~\ref{Houston_map}–\ref{DFC_map}. The qualitative visual comparison of these maps complements the quantitative results presented in Tables~\ref{Houston_acc}–\ref{DFC_acc}.

In Figs. \ref{Houston_map}(c), \ref{Berlin_map}(c) and \ref{DFC_map}(c), one can see that SVM can only identify specific general categories, such as vegetation, industrial areas, and residential zones. In contrast, when applied to HSI data, SVM exhibits an enhanced ability to discriminate a wider range of categories, confirming the expected importance of spectral information in HSI for object identification. After combining HSI-X data, SVM further refines the classification maps, which are consistent with the quantitative results in Tables~\ref{Houston_acc}–\ref{DFC_acc}.

For HSI-X data classification, we can observe that the classification maps of SVM contain a large amount of salt-pepper noise as SVM performs simple pixel-level fusion without considering spatial information. The difficulty of distinguishing some categories without spatial information is evident, especially in classes with similar spectral and elevation information (e.g., evergreen trees and deciduous trees on the DFC2018 dataset). S2FL addresses this challenge and produces smooth classification maps by incorporating spatial information and conducting feature-level fusion. After incorporating spectral and spatial features and performing adaptive cross-modality feature fusion, the patch-based networks, such as FusAtNet and Fusion\_HCT, achieve improved visual performance, resulting in fewer noisy pixels and smoother maps. However, the boundaries of these classification maps tend to be distorted, particularly those produced by CACL and MFT on the Houston2013 dataset. This distortion stems from the underlying assumption of patch-based classification that each pixel within a patch contributes equally to the definition of the class. Although this assumption is valid for most pixels in homogeneous regions, it proves disruptive when adjacent pixels belong to different categories \cite{r12}.

Compared with the patch-based algorithms, Fusion-FCN obtains slightly better boundary localization but still with many errors. For example, Fusion-FCN misclassifies part of roads, buildings, and vegetation under the shadow as water in Fig.~\ref{Houston_map}(k). Notably, our LoGoCAF consistently produces high-quality classification maps with well-positioned boundaries and smooth objects. Even under low illumination conditions or shadow regions, LoGoCAF still makes much accurate distinctions, whereas other techniques often result in wrong classifications. This is evident in the black rectangles of Fig.~\ref{Houston_map}, where some of the buildings, roads, and vegetation are covered in shadow.

\begin{figure}[tbp] 
\centerline{\includegraphics[width=8.5cm]{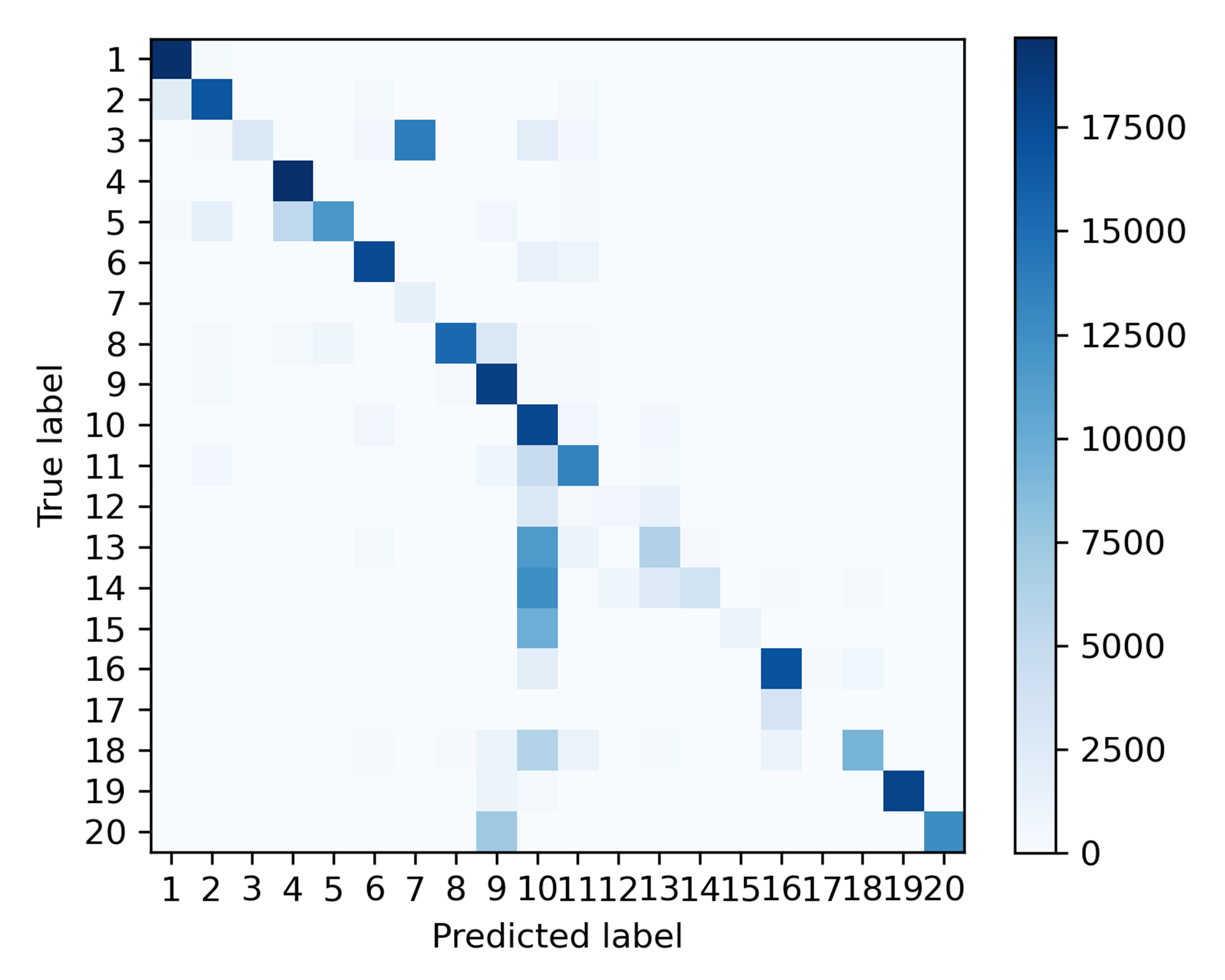}} 
\caption{The  confusion matrix of LoGoCAF on the 50-cm GSD DFC2018 dataset.}
\label{confusion} 
\end{figure}

It is worth noting that our LoGoCAF produces a more accurate classification map on the 0.5-m GSD DFC2018 dataset than on the 1-m GSD DFC2018 dataset, as shown in  Figs.~\ref{DFC_map}(k) and (l). This again emphasizes the importance of spatial information in the classification task. Although LoGoCAF achieves improved results on the 0.5-m GSD DFC2018 dataset, the confusion matrix in Fig.~\ref{confusion} reveals that LoGoCAF still incurs in many misclassifications. These misclassifications include misidentifying artificial turf as water, misidentifying crosswalks, major thoroughfares, highways, and railways as deciduous trees, and confusing stadium seats with non-residential buildings. Therefore, LoGoCAF still has room for improvement in this aspect. Overall, the qualitative investigation supports the suitability of our LoGoCAF for diverse multimodal data fusion, enabling robust semantic scene understanding.

\begin{table}[] 
\caption{Comparison of number of Params, FLOPs, Training (Train) and Inference (Infer) time of different methods on the Houston2013 dataset}
\begin{tabular}{p{5.2em} | c c c c}
\toprule[1pt]
Method         & Params     & FLOPs     & Train (s)      & Infer (s) \\ \hline
SVM            & —          & —         & 0.18          & 2.85 \\
S2FL           & —          & —         & 0.06          & 0.35 \\
FusAtNet       & 36.90M    & 221.61G    & 50583.30      & 57.15 \\
CACL           & 284.14M   & 1840.31G   & 1696081351.24 & 54.21 \\
Fusion\_HCF    & 429.84M   & 507.74G    & 2126.89       & 10.45 \\
MFT            & 313.07M   & 1619.51G   & 19010.97      & 44.79 \\
Fusion-FCN     & 94.45M    & 6215.70G   & 6720.78       & 9.60 \\ 
LoGoCAF        & 7745.97M  & 93682.68G  & 9444.23       & 22.63 \\
\bottomrule[1pt]
\end{tabular}
\label{TabComplexity}
\end{table}

\subsubsection{Model Complexity and Computation Load Evaluation}
To comprehensively analyze the efficiency of our LoGoCAF, we compared it with other methods in terms of the number of parameters (Params), floating point operations (FLOPs), training and inference time on the Houston2013 dataset.

As indicated in Table~\ref{TabComplexity}, the proposed LoGoCAF has more Params and FLOPs than other DL-based methods but exhibits relatively fast training and inference speed. Traditional methods, especially S2FL, outperform DL-based methods in training and inference efficiency. Among these patch-based models (i.e., FusAtNet, CACL, Fusion\_HCF, and MFT), Fusion\_HCT has the largest number of Params but the shortest training and inference time. Conversely, FusAtNet has the smallest number of Params but the longest inference time, while CACL has the largest FLOPs, the longest training time, and inference time comparable to Fusion\_HCT.

In general, semantic segmentation networks (i.e., Fusion-FCN and LoGoCAF) require more FLOPs and take longer training time but shorter inference time than the patch-based methods, indicating their suitability for real-world applications. This is because semantic segmentation networks require high-resolution features to ensure accuracy. In contrast, patch-based methods typically infer labels from small patches. In addition, the patch-based methods suffer from more redundant computation than the semantic segmentation networks, since the patch-based methods first generate overlapping patches and then assign labels to the corresponding central pixels to obtain a complete classification map. This severely limits their inference speed under the patch-based local learning framework.

Although our LoGoCAF does not outperform Fusion-FCN in terms of the number of Params, FLOPs, training, and inference time, its classification accuracy metrics are sharply higher than those of Fusion-FCN. This improvement is attributed to LoGoCAF's joint use of convolutions and transformers for feature extraction, along with attention-based cross-modal feature enhancement, interaction, and fusion modules (i.e., FEM and FIFM). In contrast, Fusion-FCN only uses 3 × 3 convolutions for feature extraction and a concatenate operation for feature fusion. 

Additionally, the inference speed of LoGoCAF is fast, achieving near real-time inference and demonstrating its potential for efficient multimodal data fusion and scene analysis. In short, these results demonstrate the value and practicality of our LoGoCAF.

\section{SENSITIVITY ANALYSIS}
\subsection{Model Layout Analysis}

\begin{table}[th]
 \centering
 \caption{Sensitivity analysis of the model layout in terms of OA (\%) on the Houston2013 dataset}
 \begin{tabular}{c|cccc}
 \toprule[1pt]
   Layout & C-T-T-T & C-C-T-T & C-C-C-T & C-C-C-C\\ \midrule[0.5pt]
   OA (\%)   & 87.87 &92.11 & 91.16 & 88.72  \\
\bottomrule[1pt]
\end{tabular}
\label{Tab4}
\end{table}

The layout of a model significantly affects its performance; therefore, it is necessary to explore the optimal configuration for LoGoCAF. Recognizing that convolution is more effective for processing local features in the early stages, we impose the constraint that convolution stages precede transformer stages. Under this restriction, we investigated either the convolution or the transformer block in stages 2-4. This led to four versions: C-T-T-T, C-C-T-T, C-C-C-T, and C-C-C-C, where C and T denote convolution and transformer blocks, respectively. These four layouts were then evaluated for performance.

Table \ref{Tab4} indicates that C-C-T-T achieved the best results, followed by C-C-C-T and C-T-T-T, while C-C-C-C ranked last. Therefore, we choose the C-C-T-T configuration for our LoGoCAF.

\subsection{Ablation Studies}
We conducted a series of ablation studies on the Houston2013 dataset to validate the effectiveness of each component of our LoGoCAF. 

\subsubsection{Convolution Block}
Previous studies have demonstrated that deploying convolution in shallow layers can boost performance. Our investigation explores different convolution blocks for the first two stages of our LoGoCAF, specifically Fused-MBConv without SE, Fused-MBConv with SE, and MBConv. It is worth noting that both Fused-MBConv and MBConv are the fundamental blocks of EfficientNetV2 \cite{rEfficientnetV2}. As shown in Table~\ref{Tab5}, Fused-MBConv without SE outperforms the Fused-MBConv with SE and MBConv. Therefore, the Fused-MBConv without SE is adopted.

\begin{table}[t]
 \centering
 \caption{Comparison of the performance of different convolutional blocks on the Houston2013 dataset}
 \begin{tabular}{c|cccc}
 \toprule[1pt]
   Conv Block  &  MBConv &  \makecell{Fused-MBConv \\ with SE}   & \makecell{Fused-MBConv\\ without SE} \\ \midrule[0.5pt]
     OA (\%)   & 88.56 &89.35 &  92.11 \\
\bottomrule[1pt]
\end{tabular}
\label{Tab5}
\end{table}

\subsubsection{Effectiveness of hybrid pipeline}

Firstly, we investigate the importance of the hybrid pipeline (denoted as ConvViT), where the first two stages deploy the Fused-MBConv blocks, and the last two stages use the efficient transformer blocks. Before the experiments, FEM and FIFM were removed from LoGoCAF to eliminate their interference. In addition, we set up two compared methods: one using the Fused-MBConv block in all stages (denoted as Full\_Conv) and the other using the efficient transformer block in all stages (denoted as Full\_ViT). As illustrated in Table \ref{Tab6}, Full\_Conv outperforms Full\_ViT in terms of OA by 3.18$\%$. This is consistent with previous studies that highlight the superior performance of CNN over transformer on limited training data. ConvViT significantly improves OA over Full\_Conv and Full\_ViT by 1.87$\%$ and 5.05$\%$, respectively. This suggests that hybrid architectures deploying convolutions before transformers is a compelling design choice that balances inductive biases with the representational learning capability of transformers. 

\begin{table}[]
\caption{Ablation analysis of the proposed LoGoCAF on the Houston2013 dataset}
\begin{tabular} {p{65 pt}|c c c | c}
\toprule[1pt]
Method            & Hybrid Pipeline & FEM & FIFM & OA (\%) \\ \midrule[0.5pt]
(a) Full Conv        & -           & -     & -     & 85.69 \\ 
(b) Full ViT         & -           & -     & -     & 82.51 \\   \midrule[0.5pt]
(c) ConvViT          & \ding{51}   & \ding{55}     & \ding{55}    & 87.56 \\
(d) ConvViT+FEM    & \ding{51}     & \ding{51}   & \ding{55}     & 89.26 \\
(e) ConvViT+FIFM   & \ding{51}     & \ding{55}    & \ding{51}     & 91.02 \\
(f) LoGoCAF          & \ding{51}   & \ding{51}  & \ding{51} & 92.11 \\ 
\bottomrule[1pt]   
\end{tabular}
\label{Tab6}
\end{table}

\subsubsection{Effectiveness of FEM and FIFM}
To evaluate the effectiveness of the proposed FEM and FIFM modules, we remove each module from LoGoCAF individually. Compared to ConvViT, using only FEM enhances OA by 1.70$\%$, using only FIFM improves OA by 3.46$\%$, and the combined use of FEM and FIFM leads to performance gain of 4.55$\%$. These improvements highlight the importance of the proposed FEM and FIFM in cross-model fusion and classification. 

\subsection{Impact of the Number of Training Samples}
\begin{figure}[htbp] 
\centerline{\includegraphics[width=5.5cm]{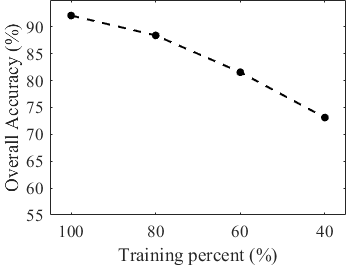}} 
\caption{Classification accuracy of LoGoCAF  with different numbers of training samples on the Houston2013 dataset.}
\label{FigNumTrn} 
\end{figure}

The effectiveness of DL-based supervised methods is particularly sensitive to the number of available training samples. To investigate the sensitivity of our LoGoCAF to different numbers of training samples, we randomly selected varying proportions from the training set of the Houston2013 dataset. As shown in Fig.~\ref{FigNumTrn}, training samples were randomly selected with proportions of [100$\%$, 80$\%$, 60$\%$, 40$\%$]. It can be seen that the classification accuracy gradually decreases as the percentage of training samples decreases. However, this decrease is relatively gradual, demonstrating the stability and robustness of our LoGoCAF. This adaptability demonstrates the potential adaptability of LoGoCAF to various real-world applications, especially in scenarios where it is difficult to obtain a large annotated dataset.

\section{Conclusion}
This study introduces a comprehensive cross-modal interaction and fusion framework for HSI-X classification, called LoGoCAF, that can be generalized to different sensing modalities. LoGoCAF enhances feature representation and optimizes performance and efficiency through a hybrid hierarchical encoder coupled with a lightweight MLP decoder. The encoder captures local and high-resolution fine details by using convolutions in shallow layers, while aggregating global and low-resolution coarse features by deploying transformers in deeper layers. The lightweight All-MLP decoder efficiently aggregates information from the encoder for feature fusion and prediction, ensuring computational efficiency without compromising performance. Moreover, the two proposed cross-modality modules, FEM and FIFM, significantly improve multimodal feature enhancement, interaction, and fusion, while consuming less computational and memory resources. Extensive experiments have shown that LoGoCAF outperforms other cross-modality fusion methods. Besides, the experiments also revealed that the spatial distribution of training and test samples has a significant impact on the classification performance. In the future, we will extend LoGoCAF to fuse HSI with multiple X models and develop extensive benchmark datasets to advance HSI-X analysis. In addition, we will investigate the fusion of any two or three types of remote sensing data.

\section*{Acknowledgment}
The authors would like to thank the IEEE GRSS IADF and Hyperspectral Image Analysis Lab at the University of Houston for providing the Houston2013 and DFC2018 datasets.

\ifCLASSOPTIONcaptionsoff
  \newpage
\fi

\bibliographystyle{IEEEtran}
\bibliography{ref}

\end{document}